\documentclass[11pt]{article}
\usepackage[]{naacl2021}
\usepackage{times}
\usepackage{latexsym}

\usepackage{microtype}

\usepackage[utf8]{inputenc} 
\usepackage[T1]{fontenc}    
\usepackage{booktabs}       
\usepackage{amsfonts}       
\usepackage{nicefrac}       
\usepackage{microtype}      
\usepackage{multirow}
\usepackage{caption}
\usepackage{amsmath,amssymb}
\usepackage{dsfont}			
\usepackage{cases}
\usepackage{color}
\usepackage{tikz}
\usepackage{pgfplots, pgfplotstable}
\usetikzlibrary{shapes,arrows,calc}
\usepackage{subfig}
\usepackage{caption}
\usepackage{textcomp}
\usepackage{enumitem}
\usepackage{import}
\usepackage{CJKutf8}
\usepackage[toc,page]{appendix}
\usepackage[textsize=large]{todonotes}
\usepackage[normalem]{ulem}
\usepackage[scaled=.8]{beramono}
\usepackage{fancyvrb}
\useunder{\uline}{\ul}{}

\setlist[itemize]{leftmargin=15pt}

\definecolor{applegreen}{rgb}{0.67, 0.88, 0.69}
\definecolor{bananamania}{rgb}{0.87, 0.72, 0.53}
\definecolor{white}{rgb}{1.0, 1.0, 1.0}
\definecolor{royalfuchsia}{rgb}{0.79, 0.17, 0.57}

\newcommand{\minus}{\scalebox{0.75}[1.0]{$-$}}

\title{Evaluating Saliency Methods for Neural Language Models}

\author{
  Shuoyang Ding\quad\quad Philipp Koehn \\
  Center for Language and Speech Processing \\
  Johns Hopkins University \\
  \texttt{\{dings, phi\}@jhu.edu}
}

\date{}

\begin{document}
\maketitle
\begin{abstract}
    Saliency methods are widely used to interpret neural network predictions, but different variants of saliency methods often disagree even on the interpretations of the same prediction made by the same model.
    In these cases, how do we identify when are these interpretations trustworthy enough to be used in analyses?
    To address this question, we conduct a comprehensive and quantitative evaluation of saliency methods on a fundamental category of NLP models: neural language models.
    We evaluate the quality of prediction interpretations from two perspectives that each represents a desirable property of these interpretations: \emph{plausibility} and \emph{faithfulness}.
    Our evaluation is conducted on four different datasets constructed from the existing human annotation of syntactic and semantic agreements, on both sentence-level and document-level.
    Through our evaluation, we identified various ways saliency methods could yield interpretations of low quality. We recommend that future work deploying such methods to neural language models should carefully validate their interpretations before drawing insights.
\end{abstract}

\section{Introduction}

While neural network models for Natural Language Processing (NLP) have recently become popular, a general complaint is that their internal decision mechanisms are hard to understand.
To alleviate this problem, recent work has deployed interpretation methods on top of the neural network models.
Among them, there is a category of interpretation methods called saliency method that is especially widely adopted~\cite{li-etal-2016-visualizing,DBLP:journals/corr/LiMJ16a,arras-etal-2016-explaining,arras-etal-2017-explaining,mudrakarta-etal-2018-model,ding-etal-2019-saliency}.
At a very high level, these methods assign an importance score to each feature in the input feature set $F$, regarding a specific prediction $y$ made by a neural network model $M$.
Such feature importance scores can hopefully shed light on the neural network models' internal decision mechanism.

\begin{table}[t]
    \hspace{-0.3cm}
    \scalebox{0.75}{
    \begin{tabular}{lp{0.6\textwidth}}
    \toprule
{\bf V} &
{\tt
\colorbox{bananamania!100.0!white}{U.S.}
\colorbox{applegreen!17.875666256231906!white}{companies}
\colorbox{bananamania!19.25966674696959!white}{wanting}
\colorbox{bananamania!7.307776201381369!white}{to}
\colorbox{bananamania!3.321797235295597!white}{expand}
\colorbox{applegreen!0.9715407857504049!white}{in}
\colorbox{bananamania!91.11381487981032!white}{Europe}
} \\
{\bf SG} &
{\tt
\colorbox{bananamania!85.2270373142!white}{U.S.}
\colorbox{applegreen!71.1828633956!white}{companies}
\colorbox{applegreen!3.5868207923!white}{wanting}
\colorbox{applegreen!3.99166149881!white}{to}
\colorbox{applegreen!6.02394457813!white}{expand}
\colorbox{applegreen!12.6632794753!white}{in}
\colorbox{bananamania!100.0!white}{Europe}
} \\
{\bf IG} &
{\tt
\colorbox{bananamania!3.0600379655608645!white}{U.S.}
\colorbox{applegreen!99.72392087724351!white}{companies}
\colorbox{applegreen!43.61218487453583!white}{wanting}
\colorbox{applegreen!100.0!white}{to}
\colorbox{applegreen!83.8267055862714!white}{expand}
\colorbox{applegreen!28.040422771537717!white}{in}
\colorbox{bananamania!82.36766101904517!white}{Europe}
} \\
\bottomrule
    \end{tabular}
    }
    \caption{An example from our evaluation where different saliency methods assign different importance scores for the same model (Transformer language model) and the same next word prediction ({\it are}).
    V, SG and IG are different saliency methods (see Section \ref{sec:saliency}).
    The tints of {\setlength{\fboxsep}{1pt}\colorbox{applegreen}{green}} and {\setlength{\fboxsep}{1pt}\colorbox{bananamania}{yellow}} mark the magnitude of positive and negative importance scores, respectively.
    }
    \label{tab:pitch}
\end{table}

While analyzing saliency interpretations uncovers useful insights for their respective task of interest, different saliency methods often give different interpretations even when the internal decision mechanism remains the same (with $F$, $y$ and $M$ held constant), as exemplified in Table \ref{tab:pitch}.
Even so, most existing work that deploys these methods often makes an ungrounded assumption that a specific saliency method can reliably uncover the internal model decision mechanism or, at most, relies merely on \emph{qualitative} inspection to determine their applicability.
Such practice has been pointed out in \citet{DBLP:conf/nips/AdebayoGMGHK18,DBLP:journals/cacm/Lipton18,belinkov-glass-2019-analysis} to be potentially problematic for model interpretation studies -- it can lead to misleading conclusions about the deep learning model's reasoning process.
On the other hand, in the context of NLP, the \emph{quantitative} evaluation of saliency interpretations largely remains an open problem~\cite{belinkov-glass-2019-analysis}.

In this paper, we address this problem by building a comprehensive quantitative benchmark to evaluate saliency methods.
Our benchmark focuses on a fundamental category of NLP models: neural language models.
Following the concepts proposed by \citet{jacovi-goldberg-2020-towards}, our benchmark evaluates the credibility of saliency interpretations from two aspects: \emph{plausibility} and \emph{faithfulness}.
In short, plausibility measures how much these interpretations align with basic human intuitions about the model decision mechanism,
while faithfulness measures how consistent the interpretations are regarding perturbations that are supposed to preserve the same model decision mechanism on either the input feature $F$ or the model $M$.

With these concepts in mind, our main contribution is materializing these tests' procedure in the context of neural language modeling and building four test sets from existing linguistic annotations to conduct these tests.
Our study covering SOTA-level models on three different network architectures reveals that saliency methods' applicability depends heavily on specific choices of saliency methods, model architectures, and model configurations. We suggest that future work deploying these methods to NLP models should carefully validate their interpretations before drawing conclusions.

This paper is organized as follows: Section 2 briefly introduces saliency methods; Section 3 describes the plausibility and faithfulness tests in our evaluation; Section 4 presents the datasets we built for the evaluation; Section 5 presents our experiment setup and results; Section 6 discusses some limitations and implications of the evaluation; Section 7 concludes the paper.

\section{Saliency} \label{sec:saliency}

The notion of \emph{saliency} discussed in this paper is a category of neural network interpretation methods that interpret a specific prediction $y$ made by a neural network model $M$, by assigning a distribution of importance $\Psi(F)$ over the input feature set $F$ of the original neural network model.

The most basic and widely used method is to assign importance by the gradient~\cite{DBLP:journals/corr/SimonyanVZ13}, which we refer to as vanilla gradient method (V). For each $x\in F$, $\psi(x) = \frac{\partial p_y}{\partial x}$, while $p_y$ is the score of prediction $y$ generated by $M$.
We also examine two improved version of gradient-based saliency: SmoothGrad (SG)~\cite{DBLP:journals/corr/SmilkovTKVW17} and Integrated Gradients (IG)~\cite{DBLP:conf/icml/SundararajanTY17}.
SmoothGrad reduces the noise in vanilla gradient-based scores by constructing several corrupted instances of the original input by adding Gaussian noise, followed by averaging the scores.
Integrated Gradients computes feature importance by computing a line integral of the vanilla saliency from a ``baseline'' point $F_0$ to the input $F$ in the feature space.
We refer the readers to the cited papers for details of these saliency methods.

There is a slight complication in the meaning of $F$ when applying these methods in the context of NLP: all the methods above will generate one importance score for each dimension of the word embedding, but most applications of saliency to NLP want a word-level importance score.
Hence, we need composition schemes to combine scores over word embedding dimensions into a single score for each word.
In the rest of this paper, we assume the ``features'' in the feature set $F$ are input words to the language model, and word-level importance scores are composed using the gradient $\cdot$ input scheme \cite{DBLP:journals/corr/DenilDF14,ding-etal-2019-saliency}.\footnote{We also experimented with the vector norm \citet{li-etal-2016-visualizing} scheme in our preliminary study, and we find it performing much worse. See details in Appendix \ref{sec:vn}.}

\section{Evaluation Paradigm}\label{sec:test-formulation}

In this section, we first introduce the notion of plausibility and faithfulness in the context of neural network interpretations (following \citet{jacovi-goldberg-2020-towards}), and then, respectively, introduce the test we adopt to evaluate them.

\subsection{Plausibility}

\noindent \textbf{Concept}\quad
An interpretation is plausible if it aligns with human intuitions about how a specific neural model makes decisions.
For example, intuitively, an image classifier can identify the object in the image because it can capture some features of the main object in the image.
Hence, a plausible interpretation would assign high importance to the area occupied by the main object.
This idea of comparison with human-annotated ground-truth (often as ``bounding-boxes'' signaling the main object's area) is used by various early studies in computer vision to evaluate saliency methods' reliability \cite[][\textit{inter alia}]{DBLP:conf/cvpr/JiangWYWZL13}.
However, the critical challenge of such evaluations for neural language models is the lack of such ground-truth annotations.

\vspace{0.1cm}
\noindent \textbf{Test}\quad
To overcome this challenge, we follow \citet{poerner-etal-2018-evaluating} to construct ground-truth annotations from existing lexical agreement annotations.
Consider, for example, the case of morphological number agreement.
Intuitively, when the language model predicts a verb with a singular morphological number, the singular nouns in the prefix should be considered important features, and vice versa.
Based on this intuition, we divide the nouns in the prefix into two different sets: the \emph{cue} set $\mathcal{C}$, which shares the same morphological number as the verb in the sentence; and the \emph{attractor} set $\mathcal{A}$, which has a different morphological number than the verb in the sentence.

Then, according to the prediction $y$ made by the model $M$, the test will be conducted under one of the two following scenarios:
\begin{itemize}[leftmargin=*,topsep=3pt] \itemsep -2pt
\item {\it Expected}: when $y$ is the verb with the correct number, the interpretation passes the test if $\max_{w\in\mathcal{C}} \psi(w) > \max_{w\in\mathcal{A}} \psi({w})$
\item {\it Alternative}: when $y$ is the verb with the incorrect number, the interpretation passes the test if $\max_{w\in\mathcal{C}} \psi(w) < \max_{w\in\mathcal{A}} \psi({w})$
\end{itemize}

However, this test has a flaw: while the evaluation criteria focus on a specific category of lexical agreement, the prediction of a word could depend on multiple lexical agreements simultaneously.
To illustrate this point, consider the verb prediction following the prefix {\it ``At the polling station people ...''}.
Suppose the model $M$ predicts the verb {\it vote}.
One could argue that {\it people} is more important than {\it polling station} because it needs the subject to determine the morphological number of the verb.
However, the semantic relation between {\it vote} and {\it polling station} is also important because that is what makes {\it vote} more likely than other random verbs, e.g. {\it sing}.

To minimize such discrepancy and constrain the scope of agreements used to make predictions, we draw inspiration from the previous work on representation probing and make adjustment to the model $M$ we are evaluating on~\cite{DBLP:conf/acl/TenneyDP19,DBLP:conf/iclr/TenneyXCWPMKDBD19,kim-etal-2019-probing,conneau-etal-2018-cram,DBLP:conf/iclr/AdiKBLG17,shi-etal-2016-string}.
The idea is to take a language model that is trained to predict words (e.g., \textit{vote} in the example above) and substitute the original final linear layer with a new linear layer (which we refer to as a \emph{probe}) fine-tuned to predict a binary lexical agreement tag (e.g., \texttt{PLURAL}) corresponding to the word choice.
By making this adjustment, the final layer extracts a subspace in the representation that is relevant to the prediction of particular lexical agreement during the forward computation, and reversely, filters out gradients that are irrelevant to the agreement prediction in the backward pass, creating an interpretation that is only subject to the same agreement constraints as to when the annotation for the test set is done.

Apart from the adjustment made on the model $M$ above, we also extend \citet{poerner-etal-2018-evaluating} in the other two aspects:
(1) we evaluate on one more lexical agreement: gender agreements between pronouns and referenced entities, and on both natural and synthetic datasets;
(2) instead of evaluating on small models, we evaluate on large SOTA-level models for each architecture.
We also show that evaluation results obtained on smaller models cannot be trivially extended to larger models.

\subsection{Faithfulness}

{\bf Concept}\quad
An interpretation is faithful if the feature importance it assigns is consistent with the internal decision mechanism of a model.
However, as \citet{jacovi-goldberg-2020-towards} pointed out, the notion of ``decision mechanism'' lacks a standard definition and a practical way to make comparisons.
Hence, as a proxy, we follow the working definition of faithfulness as proposed in their work, which states that an interpretation is faithful if the feature importance it assigns remains consistent with changes that should not change the internal model decision mechanism.
Among the three relevant factors for saliency methods (prediction $y$, model $M$, and input feature set $F$), we focus on consistency upon changes in model $M$ (model consistency) and input feature set $F$ (input consistency).\footnote{Although evaluating interpretation consistency over similar predictions $y$ is also possible, it is not of interest as most applications expect different interpretations for different predictions.}
Note that these two consistencies respectively correspond to assumptions 1 and 2 in the discussion of faithfulness evaluation in \citet{jacovi-goldberg-2020-towards}.
\vspace{0.1cm}

\noindent {\bf Model Consistency Test}\quad
To measure model consistency, we propose to measure the consistency between feature importance $\Psi_M(F)$ and $\Psi_{M'}(F)$, which is respectively generated from the original model $M$ and a smaller model $M'$ that is trained by distilling knowledge from $M$.
In this way, although $M$ and $M'$ have different architectures, $M'$ is trained to mimic the behavior of $M$ to the extent possible, and thus having similar underlying decision mechanisms.
\vspace{0.1cm}

\noindent {\bf Input Consistency Test}\quad
To measure input consistency, we perform substitutions in the input and measure the consistency between feature importance $\Psi(F)$ and $\Psi(F')$, where $F$ and $F'$ are input features sets before/after the substitution.
For example, the following prefix-prediction pairs should have the same feature importance distribution:
\begin{itemize}[leftmargin=*,topsep=3pt] \itemsep -2pt
\item {\it The \textbf{nun} bought the \textbf{son} a gift because (she...)}
\item {\it The \textbf{woman} bought the \textbf{boy} a gift because (she...)}
\end{itemize}

We measure consistency by Pearson correlation between pairs of importance score over the input feature set $F$ for both tests.
Also, note that although we can theoretically conduct faithfulness tests with any model $M$ and any dataset, for the simplicity of analysis and data creation, we will use the same model $M$ (with lexical agreement probes) and the same dataset as plausibility tests.

\setlength{\fboxsep}{2pt}
\begin{table*}[t]
    \hspace{-0.5cm}
    \scalebox{0.8}{
    \begin{tabular}{lp{1.15\textwidth}}

\toprule
{\bf PTB} &
{\tt \colorbox{bananamania!100!white}{U.S.}
\colorbox{bananamania!100!white}{Trade}
\colorbox{bananamania!100!white}{Representative}
\colorbox{bananamania!100!white}{Carla}
\colorbox{bananamania!100!white}{Hills}
\colorbox{bananamania!0!white}{said}
\colorbox{applegreen!0!white}{the}
\colorbox{applegreen!0!white}{first}
\colorbox{bananamania!0!white}{dispute-settlement}
\colorbox{bananamania!100!white}{panel}
\colorbox{applegreen!0!white}{set}
\colorbox{applegreen!0!white}{up}
\colorbox{applegreen!0!white}{under}
\colorbox{applegreen!0!white}{the}
\colorbox{bananamania!0!white}{U.S.-Canadian}
\colorbox{applegreen!0!white}{``}
\colorbox{bananamania!0!white}{free}
\colorbox{bananamania!100!white}{trade}
\colorbox{bananamania!0!white}{''}
\colorbox{bananamania!100!white}{agreement}
\colorbox{applegreen!0!white}{has}
\colorbox{applegreen!0!white}{ruled}
\colorbox{applegreen!0!white}{that}
\colorbox{bananamania!100!white}{Canada}
\colorbox{applegreen!0!white}{'s}
\colorbox{applegreen!100!white}{restrictions}
\colorbox{bananamania!0!white}{on}
\colorbox{applegreen!100!white}{exports}
\colorbox{applegreen!0!white}{of}
\colorbox{bananamania!100!white}{Pacific}
\colorbox{bananamania!100!white}{salmon}
\colorbox{bananamania!0!white}{and}
\colorbox{bananamania!100!white}{herring}
\colorbox{bananamania!0!white}{(PLURAL...)}
}
\\\midrule

{\bf Syneval} &
{\tt \colorbox{applegreen!0!white}{the}
\colorbox{applegreen!100!white}{consultant}
\colorbox{bananamania!0!white}{that}
\colorbox{applegreen!0!white}{loves}
\colorbox{bananamania!0!white}{the}
\colorbox{bananamania!100!white}{parents}
\colorbox{bananamania!0!white}{(SINGULAR...)}
}
\\\midrule

{\bf CoNLL} &
{\tt \colorbox{applegreen!100!white}{Israeli}
\colorbox{applegreen!100!white}{Prime}
\colorbox{applegreen!100!white}{Minister}
\colorbox{applegreen!100!white}{Ehud}
\colorbox{applegreen!100!white}{Barak}
\colorbox{bananamania!0!white}{says}
\colorbox{applegreen!100!white}{he}
\colorbox{applegreen!0!white}{is}
\colorbox{bananamania!0!white}{freezing}
\colorbox{applegreen!0!white}{tens}
\colorbox{applegreen!0!white}{of}
\colorbox{bananamania!0!white}{millions}
\colorbox{applegreen!0!white}{of}
\colorbox{bananamania!0!white}{dollars}
\colorbox{applegreen!0!white}{in}
\colorbox{bananamania!0!white}{tax}
\colorbox{bananamania!0!white}{payments}
\colorbox{applegreen!0!white}{to}
\colorbox{applegreen!0!white}{the}
\colorbox{applegreen!0!white}{Palestinian}
\colorbox{applegreen!0!white}{Authority}
\colorbox{applegreen!0!white}{.}
\colorbox{applegreen!100!white}{Mr.}
\colorbox{applegreen!100!white}{Barak}
\colorbox{bananamania!0!white}{says}
\colorbox{applegreen!100!white}{he}
\colorbox{applegreen!0!white}{is}
\colorbox{bananamania!0!white}{withholding}
\colorbox{applegreen!0!white}{the}
\colorbox{applegreen!0!white}{money}
\colorbox{applegreen!0!white}{until}
\colorbox{applegreen!0!white}{the}
\colorbox{applegreen!0!white}{Palestinians}
\colorbox{applegreen!0!white}{abide}
\colorbox{applegreen!0!white}{by}
\colorbox{applegreen!0!white}{cease}
\colorbox{bananamania!0!white}{-}
\colorbox{bananamania!0!white}{fire}
\colorbox{bananamania!0!white}{agreements}
\colorbox{bananamania!0!white}{.}
\colorbox{bananamania!0!white}{Earlier}
\colorbox{applegreen!0!white}{Thursday}
\colorbox{applegreen!100!white}{Mr.}
\colorbox{applegreen!100!white}{Barak}
\colorbox{bananamania!0!white}{ruled}
\colorbox{applegreen!0!white}{out}
\colorbox{bananamania!0!white}{an}
\colorbox{bananamania!0!white}{early}
\colorbox{bananamania!0!white}{resumption}
\colorbox{bananamania!0!white}{of}
\colorbox{bananamania!0!white}{peace}
\colorbox{bananamania!0!white}{talks}
\colorbox{applegreen!0!white}{,}
\colorbox{applegreen!0!white}{even}
\colorbox{bananamania!0!white}{with}
\colorbox{applegreen!0!white}{the}
\colorbox{applegreen!0!white}{United}
\colorbox{bananamania!0!white}{States}
\colorbox{applegreen!0!white}{acting}
\colorbox{applegreen!0!white}{as}
\colorbox{bananamania!0!white}{intermediary}
\colorbox{bananamania!0!white}{.}
\colorbox{bananamania!100!white}{Eve}
\colorbox{bananamania!100!white}{Conette}
\colorbox{applegreen!0!white}{reports}
\colorbox{bananamania!0!white}{from}
\colorbox{applegreen!0!white}{Jerusalem}
\colorbox{applegreen!0!white}{.}
\colorbox{bananamania!0!white}{Defending}
\colorbox{applegreen!0!white}{what}
\colorbox{applegreen!0!white}{(MASCULINE...)}
} \\\midrule

{\bf Winobias} &
{\tt \colorbox{bananamania!0!white}{The}
\colorbox{applegreen!100!white}{bride}
\colorbox{bananamania!0!white}{examined}
\colorbox{bananamania!0!white}{the}
\colorbox{bananamania!100!white}{son}
\colorbox{applegreen!0!white}{for}
\colorbox{applegreen!0!white}{injuries}
\colorbox{bananamania!0!white}{because}
\colorbox{bananamania!0!white}{(FEMININE...)}
} \\
\bottomrule

    \end{tabular}
    }
    \caption{Examples prefixes from the four evaluation datasets, followed by the probing tag prediction under the expected scenario. The cue and attractor sets are marked with solid {\setlength{\fboxsep}{1pt}\colorbox{applegreen}{Green}} and {\setlength{\fboxsep}{1pt}\colorbox{bananamania}{yellow}}, respectively.}
    \label{tab:data}
\end{table*}

\section{Data\footnote{More details on data filtering are in Appendix \ref{sec:data-filtering}.}}

Following the formulation in Section~\ref{sec:test-formulation}, we constructed four novel datasets for our benchmark, as exemplified in Table \ref{tab:data}.
Two of the datasets are concerned with \emph{number agreement} of a verb with its subject.
The other two are concerned with \emph{gender agreement} of a pronoun with its anteceding entity mentions.
For each lexical agreement type, we have one \emph{synthetic} dataset and one \emph{natural} dataset.
Both synthetic datasets ensure there is only one cue and one attractor for each test instance, while for natural datasets, there are often more than one.

For number agreement, our synthetic dataset is constructed from selected sections of \textbf{Syneval}, a targeted language model evaluation dataset from \citet{marvin-linzen-2018-targeted}, where the verbs and the subjects could be easily induced with heuristics.
We only use the most challenging sections where strongly interceding attractors are involved.
Our natural dataset for this task is filtered from Penn Treebank \cite[][\textbf{PTB}]{DBLP:journals/coling/MarcusSM94}, including training, development, and test.
We choose PTB because it offers not only human-annotated POS-tags necessary for benchmark construction but also dependent subjects of verbs for further analysis.

For gender agreement, our synthetic dataset comes from the unambiguous \textbf{Winobias} coreference resolution dataset used in \citet{jumelet-etal-2019-analysing}, and we only use the 1000-example subset where there is respectively one male and one female antecedent.
Because this dataset is intentionally designed such that most humans will find pronouns of either gender equally likely to follow the prefix, no such pronoun gender is considered to be ``correct''.
Hence, without loss of generality, we assign the female pronoun to be the expected case.\footnote{Note that this assumption will not change the interpretations we generate or the benchmark test conducted for interpretations, as we always interpret the argmax decision of the model, which is not affected by this assumption. It will only affect the breakdown of the result we report.}
Our natural dataset for this task is filtered from \textbf{CoNLL}-2012 shared task dataset for coreference resolution~\cite[also including training, development, and test]{DBLP:conf/conll/PradhanMXUZ12}.
The prefix of each test example covers a document-level context, which usually spans several hundred words.
\vspace{0.2cm}

\noindent {\bf Plausibility Test}\quad For number agreement, the cue set $\mathcal{C}$ is the set of all nouns that have the same morphological number as the verb. In contrast, the attractor set $\mathcal{A}$ is the set of all nouns with a different morphological number.
For gender agreement, the cue set $\mathcal{C}$ is the set of all nouns with the same gender as the pronoun, while the attractor set $\mathcal{A}$ is the set of all nouns with a different gender.
\vspace{0.2cm}

\noindent {\bf Model Consistency Test}\quad No special treatment to data is needed for this test. We conduct model consistency tests on all datasets we built.
\vspace{0.2cm}

\noindent {\bf Input Consistency Test}\quad We recognize that generating interpretation-preserving input perturbations for natural datasets is quite tricky.
Hence, unlike the model consistency test, we focus on the two synthetic datasets for faithfulness tests because they are generated from templates.
As can be seen from the examples, when the nouns in the cue/attractor set are substituted while maintaining the lexical agreement, the underlying model decision mechanism should be left unchanged; hence they can be viewed as interpretation-preserving perturbations.
We identified 24 and 254 such interpretation-preserving templates from our Syneval and Winobias dataset and generated perturbations pairs by combining the first example of each template with other examples generated from the same template.

\section{Experiments}
\subsection{Setup} \label{sec:exp-setup}

\begin{table*}[h]
\centering
\scalebox{0.8}{
\begin{tabular}{@{}l|rrr|rrr|rrrrrr@{}}
\toprule
                     & \multicolumn{6}{c|}{\textbf{Number Agreement}}                                                                                                                                                                        & \multicolumn{6}{c}{\textbf{Gender Agreement}}                                                                                                                                                                         \\ \midrule
                     & \multicolumn{3}{c|}{\textbf{PTB}}                                                                         & \multicolumn{3}{c|}{\textbf{Syneval}}                                                                     & \multicolumn{3}{c|}{\textbf{CoNLL}}                                                                        & \multicolumn{3}{c}{\textbf{Winobias}}                                                                    \\
                     & \multicolumn{1}{c}{\textbf{all}} & \multicolumn{1}{c}{\textbf{exp.}} & \multicolumn{1}{c|}{\textbf{alt.}} & \multicolumn{1}{c}{\textbf{all}} & \multicolumn{1}{c}{\textbf{exp.}} & \multicolumn{1}{c|}{\textbf{alt.}} & \multicolumn{1}{c}{\textbf{all}} & \multicolumn{1}{c}{\textbf{exp.}} & \multicolumn{1}{c|}{\textbf{alt.}}  & \multicolumn{1}{c}{\textbf{all}} & \multicolumn{1}{c}{\textbf{exp.}} & \multicolumn{1}{c}{\textbf{alt.}} \\ \midrule
Random               & -                                & 0.546                             & 0.454                              & -                                & 0.500                             & 0.500                              & -                                & 0.519                             & \multicolumn{1}{r|}{0.481}          & -                                & 0.500                             & 0.500                             \\
Nearest              & -                                & 0.502                             & 0.498                              & -                                & 0.140                             & 0.860                              & -                                & 0.00                              & \multicolumn{1}{r|}{1.00}           & -                                & 0.500                             & 0.500                             \\
                     & \multicolumn{1}{l}{}             & \multicolumn{1}{l}{}              & \multicolumn{1}{l|}{}              & \multicolumn{1}{l}{}             & \multicolumn{1}{l}{}              & \multicolumn{1}{l|}{}              & \multicolumn{1}{l}{}             & \multicolumn{1}{l}{}              & \multicolumn{1}{l|}{}               & \multicolumn{1}{l}{}             & \multicolumn{1}{l}{}              & \multicolumn{1}{l}{}              \\
\textbf{LSTM}        &                                  & (0.858)                           & (0.142)                            &                                  & (0.596)                           & (0.404)                            &                                  & (0.730)                           & \multicolumn{1}{r|}{(0.270)}        &                                  & (0.584)                           & (0.416)                           \\
V                    & 0.452                            & 0.484                             & 0.259                              & 0.304                            & 0.371                             & 0.206                              & 0.288                            & 0.266                             & \multicolumn{1}{r|}{0.348}          & 0.403                            & 0.440                             & 0.351                             \\
SG                   & 0.780                            & 0.805                             & \textbf{0.629}                     & \textbf{0.950}                   & \textbf{0.951}                    & \textbf{0.949}                     & \textbf{0.799}                   & \textbf{0.767}                    & \multicolumn{1}{r|}{\textbf{0.880}} & \textbf{0.984}                   & \textbf{0.981}                    & \textbf{0.988}                    \\
IG                   & \textbf{0.816}                   & \textbf{0.856}                    & 0.571                              & 0.888                            & 0.941                             & 0.811                              & 0.585                            & 0.561                             & \multicolumn{1}{r|}{0.652}          & 0.881                            & 0.853                             & 0.921                             \\
                     &                                  &                                   &                                    &                                  &                                   &                                    &                                  &                                   & \multicolumn{1}{r|}{}               &                                  &                                   &                                   \\
\textbf{QRNN}        &                                  & (0.818)                           & (0.182)                            &                                  & (0.558)                           & (0.442)                            &                                  & (0.712)                           & \multicolumn{1}{r|}{(0.288)}        &                                  & (0.715)                           & (0.285)                           \\
V                    & 0.463                            & 0.501                             & 0.289                              & 0.511                            & 0.536                             & 0.480                              & 0.669                            & 0.638                             & \multicolumn{1}{r|}{0.546}          & 0.242                            & 0.269                             & 0.175                             \\
SG                   & 0.575                            & 0.599                             & 0.468                              & 0.707                            & 0.692                             & 0.726                              & 0.503                            & 0.436                             & \multicolumn{1}{r|}{0.669}          & \textbf{0.790}                   & \textbf{0.801}                    & 0.761                             \\
IG                   & \textbf{0.697}                   & \textbf{0.728}                    & \textbf{0.555}                     & \textbf{0.797}                   & \textbf{0.764}                    & \textbf{0.838}                     & \textbf{0.737}                   & \textbf{0.700}                    & \multicolumn{1}{r|}{\textbf{0.828}} & 0.768                            & 0.730                             & \textbf{0.863}                    \\
                     &                                  &                                   &                                    &                                  &                                   &                                    &                                  &                                   & \multicolumn{1}{r|}{}               &                                  &                                   &                                   \\
\textbf{Transformer} &                                  & (0.919)                           & (0.081)                            &                                  & (0.594)                           & (0.406)                            &                                  & (0.761)                           & \multicolumn{1}{r|}{(0.239)}        &                                  & (0.219)                           & (0.781)                           \\
V                    & 0.551                            & 0.551                             & 0.551                              & 0.723                            & 0.785                             & 0.632                              & 0.674                            & 0.693                             & \multicolumn{1}{r|}{0.614}          & 0.781                            & 0.799                             & 0.766                             \\
SG                   & \textbf{0.842}                   & \textbf{0.851}                    & \textbf{0.737}                     & \textbf{0.895}                   & \textbf{0.879}                    & 0.920                              & \textbf{0.956}                   & \textbf{0.951}                    & \multicolumn{1}{r|}{\textbf{0.971}} & \textbf{0.994}                           & \textbf{1.00}                     & \textbf{0.992}                             \\
IG                   & 0.734                            & 0.741                             & 0.652                              & 0.849                            & 0.786                             & \textbf{0.940}                     & 0.829                            & 0.843                             & \multicolumn{1}{r|}{0.786}          & 0.806                   & 0.865                     & 0.775                    \\ \bottomrule
\end{tabular}
}
\caption{Plausibility benchmark result. Each number is the fraction of cases the interpretation passes the benchmark test, while the numbers in brackets for each architecture are the fraction of times these scenarios occur for predictions generated by the corresponding model. Results from the best interpretation method for each architecture are boldfaced. The \textit{exp.} and \textit{alt.} columns are breakdown of evaluation results into expected scenarios and alternative scenarios as defined in Section \ref{sec:test-formulation}. V, SG, IG stands for the vanilla saliency, SmoothGrad, and Integrated Gradients, respectively. \label{tab:main-res}} 
\end{table*}

\noindent \textbf{Interpretation Methods}\quad
For SmoothGrad (SG), we set sample size $N=30$ and sample variance $\sigma^2$ to be 0.15 times the L2-norm of word embedding matrix; for Integrated Gradients (IG), we use step size $N=100$.
These choices are made empirically and verified on a small held-out development set.

\vspace{0.1cm}
\noindent \textbf{Interpreted Model}\quad Our benchmark covers three different neural language model architectures, namely LSTM~\cite{DBLP:journals/neco/HochreiterS97}, QRNN~\cite{DBLP:conf/iclr/0002MXS17} and Transformer~\cite{DBLP:conf/nips/VaswaniSPUJGKP17,DBLP:conf/iclr/BaevskiA19,dai-etal-2019-transformer}.
All language models are trained on WikiText-103 dataset~\cite{DBLP:conf/iclr/MerityX0S17}.
For the first two architectures, we use the implementation as in \texttt{awd-lstm-lm} toolkit \cite{DBLP:conf/iclr/MerityKS18}.
For Transformer, we use the implementation in \texttt{fairseq} tookit \cite{ott-etal-2019-fairseq}.

For all the task-specific ``probes'', the fine-tuning is performed on examples extracted from WikiText\nobreakdash-2 training data.
A tuning example consists of an input prefix and a gold tag for the lexical agreement in both cases.
For number agreement, we first run Stanford POS Tagger \cite{toutanova-etal-2003-feature} on the data, and an example is extracted for each present tense verb and each instance of \textit{was} or \textit{were}.
For gender agreement, an example is extracted for each gendered pronoun.
During fine-tuning, we fix all the other parameters except the final linear layer.
The final layer is tuned to minimize cross-entropy, with Adam optimizer~\cite{DBLP:journals/corr/KingmaB14} and initial learning rate of $1e\minus 3$ with \texttt{ReduceLROnPlateau} scheduler.

We follow the setup for DistillBERT~\cite{DBLP:journals/corr/abs-1910-01108} for the distillation process involved during the model consistency test, which reduces the depth of models but not the width. For our LSTM (3 layers) and QRNN model (4 layers), the $M'$ we distill is one layer shallower than the original model $M$. For our transformer model (16 layers), we distill a 4-layer $M'$ largely due to memory constraints.

\subsection{Main Results}


\noindent {\bf Plausibility}\quad
According to our plausibility evaluation result, summarized in Table~\ref{tab:main-res}, both SG and IG consistently perform better than the vanilla saliency method regardless of different benchmark datasets and interpreted models.
However, the comparison between SG and IG interpretations varies depending on the model architecture and test sets.

\begin{table*}[t]
\hspace{-0.85cm}
\begin{tabular}{cc}
\begin{minipage}{.4\linewidth}
\centering
\scalebox{0.8}{
\begin{tabular}{@{}l|ll|ll@{}}
\toprule
                     &                   &                   &                   &                   \\
                     & \multicolumn{2}{c|}{\textbf{Syneval}} & \multicolumn{2}{c}{\textbf{Winobias}} \\
                     & \textbf{exp.}     & \textbf{alt.}     & \textbf{exp.}     & \textbf{alt.}     \\ \midrule
\textbf{LSTM}        &                   &                   &                   &                   \\
V                    & 0.532             & 0.533             & 0.447             & 0.447             \\
SG                   & 0.481             & 0.491             & 0.560             & 0.404             \\
IG                   & \textbf{0.736}    & \textbf{0.695}    & \textbf{0.735}    & \textbf{0.795}    \\
                     &                   &                   &                   &                   \\
\textbf{QRNN}        &                   &                   &                   &                   \\
V                    & 0.226             & 0.223             & \textbf{0.566}    & 0.566             \\
SG                   & 0.166             & 0.239             & 0.184             & 0.239             \\
IG                   & \textbf{0.448}    & \textbf{0.387}    & 0.499             & \textbf{0.622}    \\
                     &                   &                   &                   &                   \\
\textbf{Transformer} &                   &                   &                   &                   \\
V                    & 0.367             & 0.375             & 0.545             & 0.545             \\
SG                   & \textbf{0.604}    & \textbf{0.627}    & \textbf{0.775}    & \textbf{0.752}    \\
IG                   & 0.521             & 0.480             & 0.542             & 0.494             \\ \bottomrule
\end{tabular}
}
\end{minipage} &
\begin{minipage}{0.6\linewidth}
\centering
\scalebox{0.8}{
\begin{tabular}{@{}l|ll|ll|llll@{}}
\toprule
                     & \multicolumn{4}{c|}{\textbf{Number Agreement}}                            & \multicolumn{4}{c}{\textbf{Gender Agreement}}                                                \\ \midrule
                     & \multicolumn{2}{c|}{\textbf{PTB}} & \multicolumn{2}{c|}{\textbf{Syneval}} & \multicolumn{2}{c|}{\textbf{CoNLL}}                  & \multicolumn{2}{c}{\textbf{Winobias}} \\
                     & \textbf{exp.}   & \textbf{alt.}   & \textbf{exp.}     & \textbf{alt.}     & \textbf{exp.}  & \multicolumn{1}{l|}{\textbf{alt.}}  & \textbf{exp.}     & \textbf{alt.}     \\ \midrule
\textbf{LSTM}        &                 &                 &                   &                   &                & \multicolumn{1}{l|}{}               &                   &                   \\
V                    & 0.325           & 0.324           & 0.370             & 0.370             & \textbf{0.301} & \multicolumn{1}{l|}{\textbf{0.301}} & 0.082             & 0.082             \\
SG                   & 0.242           & 0.294           & \textbf{0.453}    & 0.394             & 0.190          & \multicolumn{1}{l|}{0.235}          & 0.071             & 0.138             \\
IG                   & \textbf{0.548}  & \textbf{0.487}  & 0.439             & \textbf{0.513}    & 0.256          & \multicolumn{1}{l|}{0.275}          & \textbf{0.435}    & \textbf{0.252}    \\
                     &                 &                 &                   &                   &                & \multicolumn{1}{l|}{}               &                   &                   \\
\textbf{QRNN}        &                 &                 &                   &                   &                & \multicolumn{1}{l|}{}               &                   &                   \\
V                    & 0.208           & 0.207           & 0.228             & 0.229             & 0.147          & \multicolumn{1}{l|}{0.147}          & 0.212             & 0.212             \\
SG                   & 0.043           & 0.044           & 0.144             & 0.131             & 0.010          & \multicolumn{1}{l|}{0.016}          & 0.063             & 0.070             \\
IG                   & \textbf{0.259}  & \textbf{0.387}  & \textbf{0.316}    & \textbf{0.350}    & \textbf{0.305} & \multicolumn{1}{l|}{\textbf{0.375}} & \textbf{0.303}    & \textbf{0.285}    \\
                     &                 &                 &                   &                   &                & \multicolumn{1}{l|}{}               &                   &                   \\
\textbf{Transformer} &                 &                 &                   &                   &                & \multicolumn{1}{l|}{}               &                   &                   \\
V                    & 0.160           & 0.160           & 0.219             & 0.219             & 0.289          & \multicolumn{1}{l|}{0.289}          & 0.104             & 0.104             \\
SG                   & \textbf{0.584}  & \textbf{0.584}  & \textbf{0.598}    & \textbf{0.570}    & \textbf{0.688} & \multicolumn{1}{l|}{\textbf{0.693}} & \textbf{0.656}    & \textbf{0.581}    \\
IG                   & 0.239           & 0.294           & 0.450             & 0.413             & 0.219          & \multicolumn{1}{l|}{0.277}          & 0.310             & 0.291             \\ \bottomrule
\end{tabular}
}
\end{minipage}\\
\rule{0pt}{3ex} (a) Input Consistency & (b) Model Consistency\\
\end{tabular}
\caption{Faithfulness Benchmark Result. Each number is the average Pearson correlation computed on the corresponding dataset. Results from the best interpretation method for each architecture are boldfaced. Refer to the caption of Table \ref{tab:main-res} for other notations. \label{tab:main-faithfulness}}
\end{table*}

Across different architectures, Transformer language model achieves the best plausibility except on the Syneval dataset.
LSTM closely follows Transformer for most benchmarks, while the plausibility of the interpretation from QRNN is much worse.
Another trend worth noting is that the gap between Transformer and the other two architectures is much larger on the CoNLL benchmark, which is the only test that involves interpreting document-level contexts.
However, these architectures' prediction accuracy is similar, meaning that there is no significant modeling power difference for gender agreements in this dataset.
We hence conjecture that the recurrent structure of LSTM and QRNN might diminish gradient signals with increasing time steps, which causes the deterioration of interpretation quality for long-distance agreements -- a problem that Transformer is exempt from, thanks to the self-attention structure.

\noindent{\bf Faithfulness}\quad
Table \ref{tab:main-faithfulness}a shows the input consistency benchmark result.
Firstly, it can be seen that the interpretations of LSTM and Transformer are more resilient to input perturbations than that of QRNN.
This is the same trend as we observed for plausibility benchmark on these datasets.
When comparing different saliency methods, we see that SG consistently outperforms for Transformer, but fails for the other two architectures, especially for QRNN.
Also, note that achieving higher plausibility does not necessarily imply higher faithfulness.
For example, compared to the vanilla saliency method, SG and IG almost always significantly improve plausibility but do not always improve faithfulness.
This lack of improvement is different from the findings in computer vision \cite{yeh2019fidelity}, where they show both SG and IG improve input consistency.
Also, for LSTM, although SG works slightly better than IG in terms of plausibility, IG outperforms SG in terms of input consistency by a large margin.

Table \ref{tab:main-faithfulness}b shows the model consistency benchmark result.
One should first notice that model consistency numbers are lower than input consistency across the board, and the drop is more significant for LSTM and QRNN even though their student model is not as different as the Transformer model ($<$20\% parameter reduction vs. 61\%).
As a result, there is a significant performance gap in terms of best model consistency results between LSTM/QRNN and Transformer.
Note that, like in plausibility results, such gap is most notable on the CoNLL dataset.
On the other hand, when comparing between saliency methods, we again see that SG outperforms for Transformer while failing most of the times for QRNN and LSTM.

\setlength{\fboxsep}{2pt}
\begin{table*}[t]
\centering
\scalebox{0.8}{
\begin{tabular}{llp{0.8\textwidth}}
\toprule
1a & QRNN+SG &
{\tt
\colorbox{applegreen!36.03984189642505!white}{The}
\colorbox{applegreen!100.0!white}{[grandmother]}
\colorbox{applegreen!26.300339196299827!white}{examined}
\colorbox{bananamania!43.926258554640945!white}{the}
\colorbox{bananamania!1.085347892306795!white}{(grandson)}
\colorbox{bananamania!38.504055869859464!white}{for}
\colorbox{applegreen!48.620851814363846!white}{injuries}
\colorbox{bananamania!11.801670113457014!white}{because}
} \\
1b & QRNN+SG &
{\tt
\colorbox{applegreen!15.93545625964086!white}{The}
\colorbox{applegreen!70.30165431467695!white}{[sister]}
\colorbox{bananamania!100.0!white}{examined}
\colorbox{bananamania!16.356945088624517!white}{the}
\colorbox{applegreen!34.34422810035785!white}{(groom)}
\colorbox{applegreen!8.632503512915275!white}{for}
\colorbox{bananamania!0.7378140921992494!white}{injuries}
\colorbox{applegreen!9.516095017353999!white}{because}
} \\
\midrule
2a & QRNN+V &
{\tt
\colorbox{applegreen!61.40635302792366!white}{The}
\colorbox{bananamania!99.0692028116589!white}{[grandmother]}
\colorbox{bananamania!39.6128316451768!white}{examined}
\colorbox{bananamania!90.02944541856829!white}{the}
\colorbox{applegreen!100.0!white}{(grandson)}
\colorbox{bananamania!51.35414020193972!white}{for}
\colorbox{applegreen!90.05607584400735!white}{injuries}
\colorbox{bananamania!13.465819627535252!white}{because}
} \\
2b & QRNN+V &
{\tt
\colorbox{applegreen!35.33466003180534!white}{The}
\colorbox{bananamania!100.0!white}{[aunt]}
\colorbox{bananamania!52.75079882332301!white}{examined}
\colorbox{bananamania!19.660005495539977!white}{the}
\colorbox{applegreen!29.130917956670455!white}{(groom)}
\colorbox{bananamania!23.025436572651692!white}{for}
\colorbox{applegreen!32.167505393188435!white}{injuries}
\colorbox{bananamania!11.48266094250949!white}{because}
} \\
\midrule
3a & QRNN+SG &
{\tt
\colorbox{applegreen!36.03984189642505!white}{The}
\colorbox{applegreen!100.0!white}{[grandmother]}
\colorbox{applegreen!26.300339196299827!white}{examined}
\colorbox{bananamania!43.926258554640945!white}{the}
\colorbox{bananamania!1.085347892306795!white}{(grandson)}
\colorbox{bananamania!38.504055869859464!white}{for}
\colorbox{applegreen!48.620851814363846!white}{injuries}
\colorbox{bananamania!11.801670113457014!white}{because}
} \\
3b & QRNN\_distilled+SG &
{\tt
\colorbox{applegreen!100.0!white}{The}
\colorbox{bananamania!72.35982423000866!white}{[grandmother]}
\colorbox{applegreen!70.75793462641984!white}{examined}
\colorbox{bananamania!27.249387551209715!white}{the}
\colorbox{bananamania!4.250832651020821!white}{(grandson)}
\colorbox{bananamania!16.828251692237817!white}{for}
\colorbox{applegreen!81.70612833327952!white}{injuries}
\colorbox{bananamania!30.63489345585816!white}{because}
} \\
\midrule
4a & Transformer+SG &
{\tt
\colorbox{applegreen!19.809388416576887!white}{The}
\colorbox{applegreen!100.0!white}{[grandmother]}
\colorbox{bananamania!29.886476664775373!white}{examined}
\colorbox{applegreen!3.5512185898481383!white}{the}
\colorbox{bananamania!26.366345913091017!white}{(grandson)}
\colorbox{applegreen!39.26892649007991!white}{for}
\colorbox{bananamania!2.077600360876487!white}{injuries}
\colorbox{applegreen!0.3639404631683738!white}{because}
} \\
4b & Transformer+SG &
{\tt
\colorbox{applegreen!27.378843444362605!white}{The}
\colorbox{applegreen!100.0!white}{[aunt]}
\colorbox{bananamania!50.52628850501851!white}{examined}
\colorbox{bananamania!80.53278652763841!white}{the}
\colorbox{applegreen!3.9358365177579255!white}{(groom)}
\colorbox{applegreen!25.526214080588826!white}{for}
\colorbox{bananamania!5.26443769777009!white}{injuries}
\colorbox{applegreen!33.21603615487467!white}{because}
} \\
4c & Transformer\_distilled+SG &
{\tt
\colorbox{bananamania!35.390838534764626!white}{The}
\colorbox{applegreen!100.0!white}{[grandmother]}
\colorbox{bananamania!7.592479590430636!white}{examined}
\colorbox{bananamania!26.007180138828957!white}{the}
\colorbox{bananamania!9.29268098157438!white}{(grandson)}
\colorbox{applegreen!18.53452793366556!white}{for}
\colorbox{applegreen!0.36581842710743134!white}{injuries}
\colorbox{applegreen!6.341937913201518!white}{because}
} \\
\bottomrule
\end{tabular}
}
\caption{Examples from Winobias dataset for qualitative analysis. Cue words are marked with [] while attractor words are marked with (). The tints of {\setlength{\fboxsep}{1pt}\colorbox{applegreen}{green}} and {\setlength{\fboxsep}{1pt}\colorbox{bananamania}{yellow}} mark the magnitude of positive and negative importance scores, respectively. For all examples, the prediction interpreted is the {\tt FEMININE} tag. 1 is a case with high plausibility and low input faithfulness; 2 is a case with low plausibility and high input faithfulness; 3 is a case with low model faithfulness; 4 is a case with high plausibility and high input/model faithfulness. \label{tab:gen-exp}}
\end{table*}

\subsection{Analysis} \label{sec:analysis}

\vspace{0.1cm}
\noindent{\bf Plausibility vs. Faithfulness}\quad
A natural question for our evaluation is how the property of plausibility and faithfulness interact with each other.
Table~\ref{tab:gen-exp} illustrates such interaction with qualitative examples.
Among them, 1 and 2 are two cases where the plausibility and input faithfulness evaluation results do not correlate.
In general, the interpretations in both cases are of low quality, but they also fail in different ways.
In case 1, the interpretation assigns the correct relative ranking for the cue words and attractor words, but the importance of the words outside the cue/attractor set varies upon perturbation.
On the other hand, in case 2, the importance ranking among features is roughly maintained upon perturbation, but the importance score assigned for both examples do not agree with the prediction interpreted ({\tt FEMININE} tag) and thus can hardly be understood by humans.
It should be noted that these defects can only be revealed when both plausibility and faithfulness tests for interpretations are deployed.

Case 3 shows a scenario where the saliency method yields very different interpretations for the same input/prediction pair, indicating that interpretations from this architecture/saliency method combination are subject to changes upon changes in the architecture configurations.
Finally, in case 4, we see that an architecture/saliency method combination performing well in all tests yields stable interpretations that humans can easily understand.



\begin{table}[h]
\hspace{-0.25cm}
\scalebox{0.75}{
\begin{tabular}{@{}l|lll|lll@{}}
\toprule
                                                                             & \multicolumn{3}{c|}{\textbf{Syneval}}        & \multicolumn{3}{c}{\textbf{Winobias}}        \\
                                                                             & \textbf{all} & \textbf{exp.} & \textbf{alt.} & \textbf{all} & \textbf{exp.} & \textbf{alt.} \\ \midrule
\textbf{best plausibility}                                                   &              &               &               &              &               &               \\
LSTM (SG)                                                                    & 0.945        & 0.922         & 0.973         & 0.948        & 0.950         & 0.904         \\
QRNN (IG)                                                                    & 0.981        & 0.964         & 0.998         & 0.974        & 0.974         & 1.00          \\
Transformer (SG)                                                             & 0.917        & 0.908         & 0.929         & 0.997        & 1.00          & 0.996         \\
& & & & & & \\
\textbf{\begin{tabular}[c]{@{}l@{}}best (input)\\ faithfulness\end{tabular}} &              &               &               &              &               &               \\
LSTM (IG)                                                                    & --           & 0.628         & 0.739         & --           & 0.820         & 0.769         \\
QRNN (IG)                                                                    & --           & 0.733         & 0.831         & --           & 0.891         & 0.841         \\
Transformer (SG)                                                             & --           & 0.569         & 0.581         & --           & 0.932         & 0.912         \\ \bottomrule
\end{tabular}
}
\caption{Plausibility \& input faithfulness on synthetic datasets with distilled models. Only results for the interpretation method with best performance are shown. Refer to the caption of Table \ref{tab:main-res} for other notations. \label{tab:distilled}}
\end{table}

\vspace{0.1cm}
\noindent{\bf Sensitivity to Model Configurations}\quad
Our model faithfulness evaluation shows that variations in the model configurations (number of layers) could drastically change the model interpretation in many cases.
Hence, we want to answer two analysis questions: (1) are these interpretations changing for the better or worse quality-wise with the distilled smaller models? (2) are there any patterns for such changes?
Due to space constraints, we only show some analysis results for question (1) in Table \ref{tab:distilled}.
Overall, compared to the corresponding results in Table~\ref{tab:main-res} (for plausibility) and Table~\ref{tab:main-faithfulness}a (for input faithfulness), the saliency methods we evaluated perform better with the smaller distilled models.
Most remarkably, we see a drastic performance improvement for QRNN, both in plausibility and faithfulness.
For LSTM and Transformer, we observe an improvement for input faithfulness on Winobias and roughly the same performance for other tests.

As for the second question, we build smaller Transformer language models with various depth, number of heads, embedding size, and feed-forward layer width settings, while keeping other hyperparameters unchanged.
Unfortunately, the trends are quite noisy and also heavily depends on the chosen saliency methods.\footnote{Detailed discussion of these analyses is in Appendix \ref{sec:transformer-analysis}.}
Hence, it is highly recommended that evaluation of saliency methods be conducted on the specific model configurations of interest, and trends of interpretation quality on a specific model configuration should not be over-generalized to other configurations.

\begin{table}[t]
    \centering
    \scalebox{0.75}{
    \begin{tabular}{p{0.6\textwidth}}
    \toprule
{\bf V} \quad
{\tt
\colorbox{bananamania!5.076405660849186!white}{``}
\colorbox{applegreen!3.143233572620151!white}{The}
\colorbox{applegreen!0.7170318214869243!white}{[fact]}
\colorbox{bananamania!30.491838136480787!white}{that}
\colorbox{applegreen!1.158188838404898!white}{this}
\colorbox{applegreen!14.775880205464878!white}{happened}
\colorbox{applegreen!2.8554055865683714!white}{two}
\colorbox{bananamania!8.089601566289414!white}{(years)}
\colorbox{applegreen!16.086394654859077!white}{ago}
\colorbox{bananamania!3.8290520998535396!white}{and}
\colorbox{bananamania!35.478489045355396!white}{there}
\colorbox{bananamania!0.4386290461958604!white}{was}
\colorbox{bananamania!14.926924028482402!white}{a}
\colorbox{applegreen!100.0!white}{[recovery]}
} \\\midrule
{\bf SG}
{\tt
\colorbox{bananamania!1.436785045146077!white}{``}
\colorbox{applegreen!10.179388951296827!white}{The}
\colorbox{applegreen!8.118193473587285!white}{[fact]}
\colorbox{applegreen!4.720226316501814!white}{that}
\colorbox{bananamania!1.6133057651690461!white}{this}
\colorbox{applegreen!12.741736511437956!white}{happened}
\colorbox{applegreen!2.56109462979424!white}{two}
\colorbox{applegreen!5.9121583039504255!white}{(years)}
\colorbox{applegreen!21.005044566567268!white}{ago}
\colorbox{applegreen!6.260754908702671!white}{and}
\colorbox{bananamania!49.277566496859606!white}{there}
\colorbox{bananamania!25.69109608706313!white}{was}
\colorbox{applegreen!14.298695951982612!white}{a}
\colorbox{applegreen!100.0!white}{[recovery]}
} \\\midrule
{\bf IG} \quad
{\tt
\colorbox{bananamania!100.0!white}{``}
\colorbox{applegreen!26.12878837011616!white}{The}
\colorbox{applegreen!9.907748989976248!white}{[fact]}
\colorbox{applegreen!4.284943273475783!white}{that}
\colorbox{applegreen!11.229258493745382!white}{this}
\colorbox{bananamania!3.8174632288049626!white}{happened}
\colorbox{bananamania!11.508882124839491!white}{two}
\colorbox{bananamania!2.8161803520084767!white}{(years)}
\colorbox{bananamania!16.950770195845774!white}{ago}
\colorbox{bananamania!8.88252049350986!white}{and}
\colorbox{bananamania!20.955843552153038!white}{there}
\colorbox{bananamania!4.832269035427887!white}{was}
\colorbox{bananamania!35.07848681106145!white}{a}
\colorbox{applegreen!39.52100642341448!white}{[recovery]}
} \\
\bottomrule
    \end{tabular}
    }
    \caption{A number agreement test case where the distilled Transformer model makes the correct prediction (singular) but all interpretation methods unanimously point to a singular noun that is not grammatical subject as the most salient cue for this prediction.}
    \label{tab:probe}
\end{table}

\vspace{0.1cm}
\noindent{\bf Saliency vs. Probing}\quad
Our evaluation incorporates probing to focus only on specific lexical agreements of interest.
It should be pointed out that in the literature of representation probing, the method has always been working under the following assumption: when the model makes an expected-scenario ("correct") prediction, it is always referring to a \emph{grammatical} cue, for example, the subject of the verb in the number agreement case.
However, in our evaluation, we also observe some interesting phenomena in the interpretation of saliency methods that breaks the assumption, which is exemplified in Table~\ref{tab:probe}.
This calls for future work that aims to better understand language model behaviors by examining other possible cues used for predictions made in representation probing under the validated cases where saliency methods could be reliably applied.

\section{Discussion}



Most existing work on evaluating saliency methods focuses only on computer vision models~\cite[][\textit{inter alia}]{DBLP:journals/corr/abs-2011-05429,DBLP:conf/nips/HookerEKK19,DBLP:conf/nips/AdebayoGMGHK18,DBLP:conf/nips/HeoJM19,DBLP:conf/aaai/GhorbaniAZ19}.
In the context of NLP, \citet{poerner-etal-2018-evaluating} is the first work to conduct such evaluations for NLP and the only prior work that conducts such evaluations for neural language models but has several limitations as we have already pointed out in Section \ref{sec:test-formulation}.
\citet{arras-etal-2019-evaluating,atanasova-etal-2020-diagnostic,hao-2020-evaluating} conducted similar evaluations based on specifically designed diagnostic toy tasks and/or text classification, while~\citet{bastings-filippova-2020-elephant} casted doubt on whether these conclusions could be generalized to sequence generation tasks.
\citet{li-etal-2020-evaluating} evaluated various interpretation methods for neural machine translation models by building proxy models on only the top-$k$ important input words as determined by the interpretation methods, but such evaluation requires generating interpretations for a large training set and hence is intractable for even mildly computationally-expensive methods such as SmoothGrad and Integrated Gradients.
On a slightly different line, \citet{deyoung-etal-2020-eraser} built a benchmark to evaluate a specific category of NLP models that generate rationales during predictions, which is a different path towards building explainable NLP models.

Our evaluation is not without its limitations.
The first limitation, inherited from earlier work by \citet{poerner-etal-2018-evaluating}, is that our plausibility test only concerns the words in cue/attractor sets rather than other words in the input prefix.
Such limitation is inevitable because the annotations from which we build our ground-truth interpretations are only concerned with a specific lexical agreement.
This limitation can be mitigated by combining plausibility tests with faithfulness tests, which concern all the input prefix words.

The second limitation is that the test sets used in these benchmarks need to be constructed in a case-to-case manner, according to the chosen lexical agreements and the input perturbations.
While it is hard to create plausibility test sets without human interference, future work could explore automatic input consistency tests by utilizing adversarial input generation techniques in NLP~\cite{alzantot-etal-2018-generating,cheng-etal-2019-robust,cheng-etal-2020-advaug}.

It should also be noted that while our work focuses on evaluating a specific category of interpretation methods for neural language models, our evaluation paradigm can be easily extended to evaluating other interpretation methods such as attention mechanism, and with other sequence models such as masked language models (e.g., BERT). 
We would also like to extend these evaluations beyond English datasets, especially to languages with richer morphological inflections.


\section{Conclusion}



We conduct a quantitative evaluation of saliency methods on neural language models based on the perspective of plausibility and faithfulness.
Our evaluation shows that a model interpretation can either fail due to a lack of plausibility or faithfulness, and the interpretations are trustworthy only when they do well with both tests.
We also noticed that the performance of saliency interpretations are generally sensitive to even minor model configuration changes.
Hence, trends of interpretation quality on a specific model configuration should not be over-generalized to other configurations.

We want the community to be aware that saliency methods, like many other post-hoc interpretation methods, still do not generate trustworthy interpretations all the time.
Hence, we recommend that adopting any model interpretation method as a source of knowledge about NLP models' reasoning process should only happen after similar quantitative checks as presented in this paper are performed.
We also hope our proposed test paradigm and accompanied test sets provide useful guidance to future work on evaluations of interpretation methods.
Our evaluation dataset and code to reproduce the analysis are available at \url{https://github.com/shuoyangd/tarsius}.

\section*{Acknowledgements}
The authors would like to thank colleagues at CLSP and anonymous reviewers for feedback at various stages of the draft. This material is based upon work supported by the United States Air Force under Contract  No. FA8750-19-C-0098. Any opinions, findings, and conclusions or recommendations expressed in this material are those of the author(s) and do not necessarily reflect the views of the United States Air Force and DARPA.

\bibliography{anthology.bib}
\bibliographystyle{acl_natbib}

\clearpage
\appendix
\section{Data Filtering Details} \label{sec:data-filtering}

\subsection{Penn Treebank (PTB)}

A potential candidate for a test case is extracted every time a word with POS tag VBZ (Verb, 3rd person singular present) or VBP (Verb, non-3rd person singular present), or a copula that is among \emph{is, are, was, were}, shows up.
The candidate will then be filtered subjecting to the following criteria:
\begin{enumerate}
\item The prefix has at least one attractor word (a noun that has different morphological number as the verb that is predicted). This is to ensure that evaluation could be conducted in alternative scenario.
\item The verb cannot immediately follow its grammatical subject (note: it may still immediately follow a cue word that is not a grammatical subject). This is to ensure that the signal of the subject is not overwhelmingly-strong compared to the attractors.
\item Not all attractors occur earlier 10 words than the grammatical subject. Same reason as the previous criteria.
\end{enumerate}

Overall, we obtained 1448 test cases out of 49168 sentences in PTB (including train, dev and test set).
We lose a vast majority of sentences mostly because of the last two criteria.

\subsection{Syneval}

We use the following sections of the original data \cite[followed by their names in the data dump, ][]{marvin-linzen-2018-targeted}:

\begin{itemize}
\item Agreement in a sentential complemenet: {\tt sent\_comp}
\item Agreement across a prepositional phrase: {\tt prep\_anim} and {\tt prep\_inanim}
\item Agreement across a subject relative clause: {\tt subj\_rel}
\item Agreement across an object relative clause: \\ {\tt obj\_rel\_across\_anim}, {\tt obj\_rel\_across\_inanim}, {\tt obj\_rel\_no\_comp\_across\_anim}, {\tt obj\_rel\_no\_comp\_across\_inanim}
\item Agreement within an object relative clause: \\ {\tt obj\_rel\_within\_anim}, {\tt obj\_rel\_within\_inanim}, {\tt obj\_rel\_no\_comp\_within\_anim}, {\tt obj\_rel\_no\_comp\_within\_inanim}
\end{itemize}

We select these sections because they all have strong interfering attractors or have cues that may potentially be mistaken as attractors.
We obtained much less examples (6280) than the original data (249760) because lots of examples only differ in the verb or the object they use, which become duplicates when we extract prefix before the verb.

The original dataset does not come with cue/attractor annotations, but it can be easily inferred because they are generated by simple heuristics.

Note that most of these sections have only around 50\% prediciton accuracy with RNNs in the original paper.
Our results on large-scale language models corroborate the findings in the original paper.

\subsection{CoNLL}

We use the dataset \cite{DBLP:conf/conll/PradhanMXUZ12} with gold parses, entities mentions and mention boundaries.
A potential candidate for a test case is extracted every time a pronoun shows up.
The male pronouns are \emph{he, him, himself, his}, while the female pronouns are \emph{she, her, herself, hers}.
We don't consider cases epicene pronouns like \emph{it, they}, etc. because they often involve tricky cases like entity mentions covering a whole clause.
We break prefixes according to the document boudaries as provided in the original dataset unless the prefix is longer than 512 words, in which case we instead break at the nearest sentence boundary.

The annotation for this dataset does not cover the gender of entities.
We are aware that the original shared task provides gender annotation, but to this day, the documentation for the data is missing and hence we cannot make use of this annotation.
Hence, we instead used several heuristics to infer the gender of an entity mention, in desending order:
\begin{itemize}
\item If an entity mention and a pronoun have coreference relationship, they should share the same gender.
\item If an entity mention starts with ``Mr.'' or ``Mrs.'' or ``Ms.'', we assign the corresponding gender.
\item If the entity mention has a length of two tokens, we assume its a name and use gender inference tools\footnote{\tt https://github.com/lead-ratings/gender-guesser} to guess its gender.
Note that the gender guesser may also indicate that it's not able to infer the gender, in that case we do not assign a gender.
\item If a mention is coreferenced with another mention that is not pronoun, they should also have the same gender.
\end{itemize}
Manual inspection of the resulting data indicates that the scheme above covers the gender of most entity mentions correctly.
We hope that our dataset could be further perfected by utilizing higher quality annotation on entity genders.

Since each entity mention could span more than one words, we add all words within the span into their corresponding cue/attractor set.
A tricky case is where two entity mention spans are nested or intersected.
For the first case, we exclude smaller span from the larger one to create two unintersected spans as the new span for cue/attracor set.
For the second case, we exclude the intersecting parts from both spans.

Finally, all candidates are filtered subjecting to the following two criteria:
\begin{enumerate}
\item The prefix should include one attractor entity.
\item The entity mention that is cloest to the verb should be an attractor.
\end{enumerate}

We obtained 586 document segments from the 2280 documents in the original data.
As pointed out in \citet{zhao-etal-2018-gender}, the CoNLL dataset is significantly biased towards male entity mentions.
Nevertheless, our filtering scheme generated a relatively balanced test set: among the 586 test cases, 258 are male pronouns, while 328 are female pronouns.

\subsection{Winobias}

We used the same data as the unambiguous coreference resolution dataset in \citet{jumelet-etal-2019-analysing}, which is in turn generated by a script from \citet{zhao-etal-2018-gender}, except that we excluded the cases where both nouns in the sentence are of the same gender.
Similar to \textbf{Syneval} dataset, the cue and attractors could easily be inferred with heuristics.

\section{Additional Results}

We leave some results that we cannot fit into the main paper here.

\subsection{Vector Norm (VN) Composition Scheme} \label{sec:vn}

\begin{table*}[h]
\scalebox{0.8}{
\begin{tabular}{@{}l|rrr|rrr|rrrrrr@{}}
\toprule
                     & \multicolumn{6}{c|}{\textbf{Number Agreement}}                                                                                                                                                                        & \multicolumn{6}{c}{\textbf{Gender Agreement}}                                                                                                                                                                        \\ \midrule
                     & \multicolumn{3}{c|}{\textbf{PTB}}                                                                         & \multicolumn{3}{c|}{\textbf{Syneval}}                                                                     & \multicolumn{3}{c|}{\textbf{CoNLL}}                                                                       & \multicolumn{3}{c}{\textbf{Winobias}}                                                                    \\
                     & \multicolumn{1}{c}{\textbf{all}} & \multicolumn{1}{c}{\textbf{exp.}} & \multicolumn{1}{c|}{\textbf{alt.}} & \multicolumn{1}{c}{\textbf{all}} & \multicolumn{1}{c}{\textbf{exp.}} & \multicolumn{1}{c|}{\textbf{alt.}} & \multicolumn{1}{c}{\textbf{all}} & \multicolumn{1}{c}{\textbf{exp.}} & \multicolumn{1}{c|}{\textbf{alt.}} & \multicolumn{1}{c}{\textbf{all}} & \multicolumn{1}{c}{\textbf{exp.}} & \multicolumn{1}{c}{\textbf{alt.}} \\ \midrule
\textbf{LSTM}        &                                  & (0.858)                           & (0.142)                            &                                  & (0.596)                           & (0.404)                            &                                  & (0.730)                           & \multicolumn{1}{r|}{(0.270)}       &                                  & (0.584)                           & (0.416)                           \\
V+VN                 & 0.683                            & 0.719                             & 0.463                              & 0.643                            & 0.466                             & 0.903                              & 0.459                            & 0.393                             & \multicolumn{1}{r|}{0.639}         & 0.680                            & 0.807                             & 0.502                             \\
SG+VN                & 0.543                            & 0.540                             & 0.561                              & 0.549                            & 0.271                             & 0.959                              & 0.394                            & 0.234                             & \multicolumn{1}{r|}{0.829}         & 0.625                            & 0.587                             & 0.678                             \\
                     &                                  &                                   &                                    &                                  &                                   &                                    &                                  &                                   & \multicolumn{1}{r|}{}              &                                  &                                   &                                   \\
\textbf{QRNN}        &                                  & (0.818)                           & (0.182)                            &                                  & (0.558)                           & (0.442)                            &                                  & (0.712)                           & \multicolumn{1}{r|}{(0.288)}       &                                  & (0.715)                           & (0.285)                           \\
V+VN                 & 0.630                            & 0.673                             & 0.437                              & 0.579                            & 0.456                             & 0.735                              & 0.427                            & 0.309                             & \multicolumn{1}{r|}{0.716}         & 0.526                            & 0.650                             & 0.214                             \\
SG+VN                & 0.559                            & 0.567                             & 0.521                              & 0.556                            & 0.352                             & 0.813                              & 0.398                            & 0.230                             & \multicolumn{1}{r|}{0.811}         & 0.539                            & 0.538                             & 0.540                             \\
                     &                                  &                                   &                                    &                                  &                                   &                                    &                                  &                                   & \multicolumn{1}{r|}{}              &                                  &                                   &                                   \\
\textbf{Transformer} &                                  & (0.919)                           & (0.081)                            &                                  & (0.594)                           & (0.406)                            &                                  & (0.761)                           & \multicolumn{1}{r|}{(0.239)}       &                                  & (0.219)                           & (0.781)                           \\
V+VN                 & 0.604                            & 0.620                             & 0.424                              & 0.671                            & 0.525                             & 0.885                              & 0.507                            & 0.511                             & \multicolumn{1}{r|}{0.493}         & 0.481                            & 0.840                             & 0.380                             \\
SG+VN                & 0.592                            & 0.596                             & 0.542                              & 0.654                            & 0.504                             & 0.872                              & 0.529                            & 0.538                             & \multicolumn{1}{r|}{0.500}         & 0.437                            & 0.836                             & 0.325                             \\ \bottomrule
\end{tabular}}
\caption{Plausibility benchmark result for Vector Norm (VN) composition scheme. Refer to the caption of Table \ref{tab:main-res} for notations. \label{tab:main-pl-vn}}
\end{table*}

In this section, we explain why we chose not to cover the vector norm composition scheme (mentioned in \ref{sec:saliency}) in our main evaluation results.

We would like to argue first that even mathematically, VN is not a good fit for our evaluation paradigm.
Vector norm composition scheme will only indicate the importance of a feature, but will not indicate the polarity of the importance because it cannot generate a negative word importance score, which is important for our evaluation.
The reason why it is important is that our plausibility evaluation does distinguish between input words that should have positive/negative importance scores by placing them in cue and attractor sets, respectively.
For example, in Table \ref{tab:pitch}, the singular proper noun \emph{U.S.} and \emph{Europe} are important input words because they could potentially lead the model to make the alternative prediction \emph{is} instead of the expected prediction \emph{are}.
Hence, they are placed into the attractor set, and when interpreting the next word prediction \emph{are}, our plausibility test expects that they should have large negative importance scores.

Besides, we did run the plausibility evaluation with vector norm composition scheme under some settings, as shown in Table \ref{tab:main-pl-vn}.
For the vanilla gradient saliency method, the VN composition scheme performs on-par with the gradient $\cdot$ input (GI) scheme (which is used for our main results).
However, with SmoothGrad, the plausibility result does not significantly change like the case with the gradient $\cdot$ input (GI) scheme.
This corroborates with the results in \cite{ding-etal-2019-saliency}, where they also show that SmoothGrad does not improve the interpretation quality with VN composition scheme.

With these theoretical and empirical evidence, we decided to drop vector norm composition scheme for our evaluation.

\subsection{Patterns for Changes of Interpretation Quality with Varying Model Configurations} \label{sec:transformer-analysis}

\begin{figure*}[h]
\includegraphics[width=\textwidth]{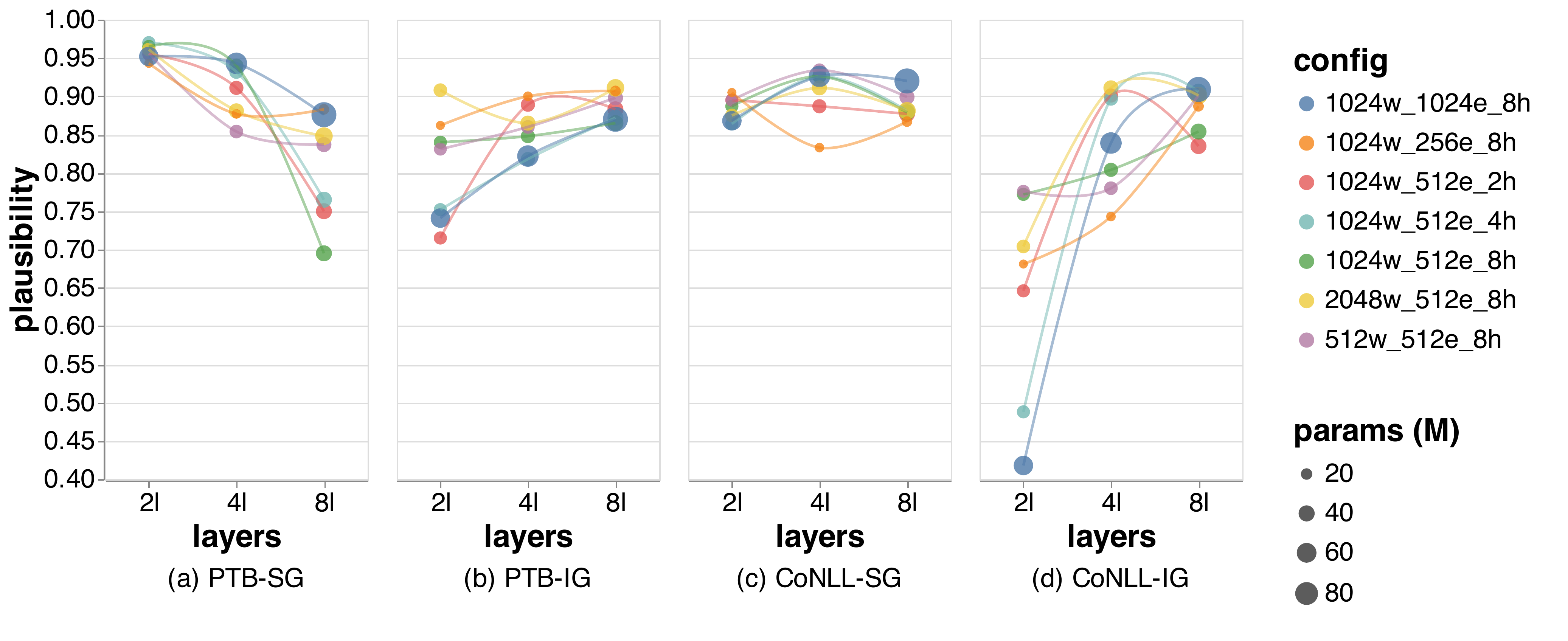}
\caption{Analysis of model configuration vs. plausibility on PTB and CoNLL benchmark. Each model configuration is color-coded, while the parameter size (in millions) is shown with circle size. \texttt{l}, \texttt{w}, \texttt{e}, \texttt{h} stands for model depth, width of feed-forward layers after self-attention, embedding size, and the number of heads. \label{fig:config-sg-all}}
\end{figure*}

\begin{figure*}[h]
\centering
\begin{tabular}{cc}
\includegraphics[width=0.45\textwidth]{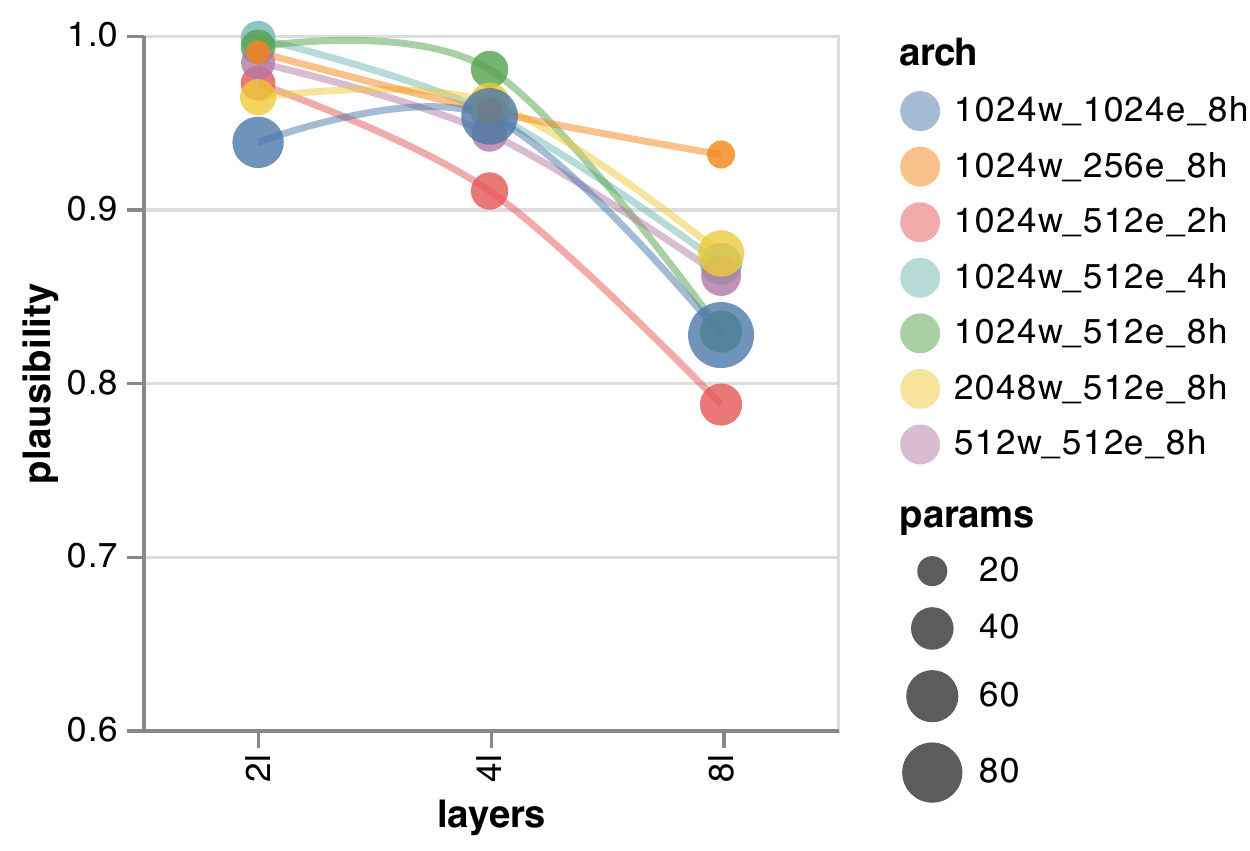} &
\includegraphics[width=0.45\textwidth]{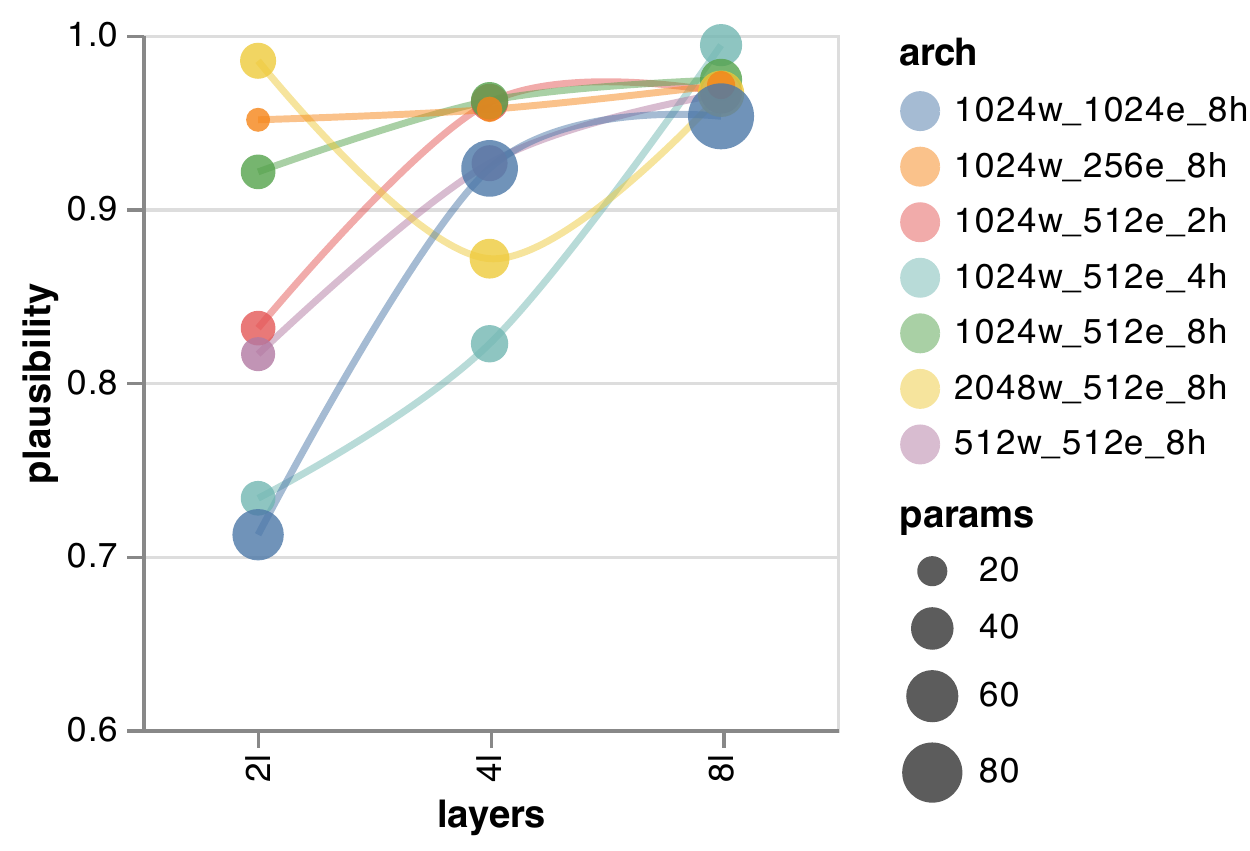} \\
(a) Plausibility benchmark with SmoothGrad & (b) Plausibility benchmark with Integrated Gradients \\
\includegraphics[width=0.45\textwidth]{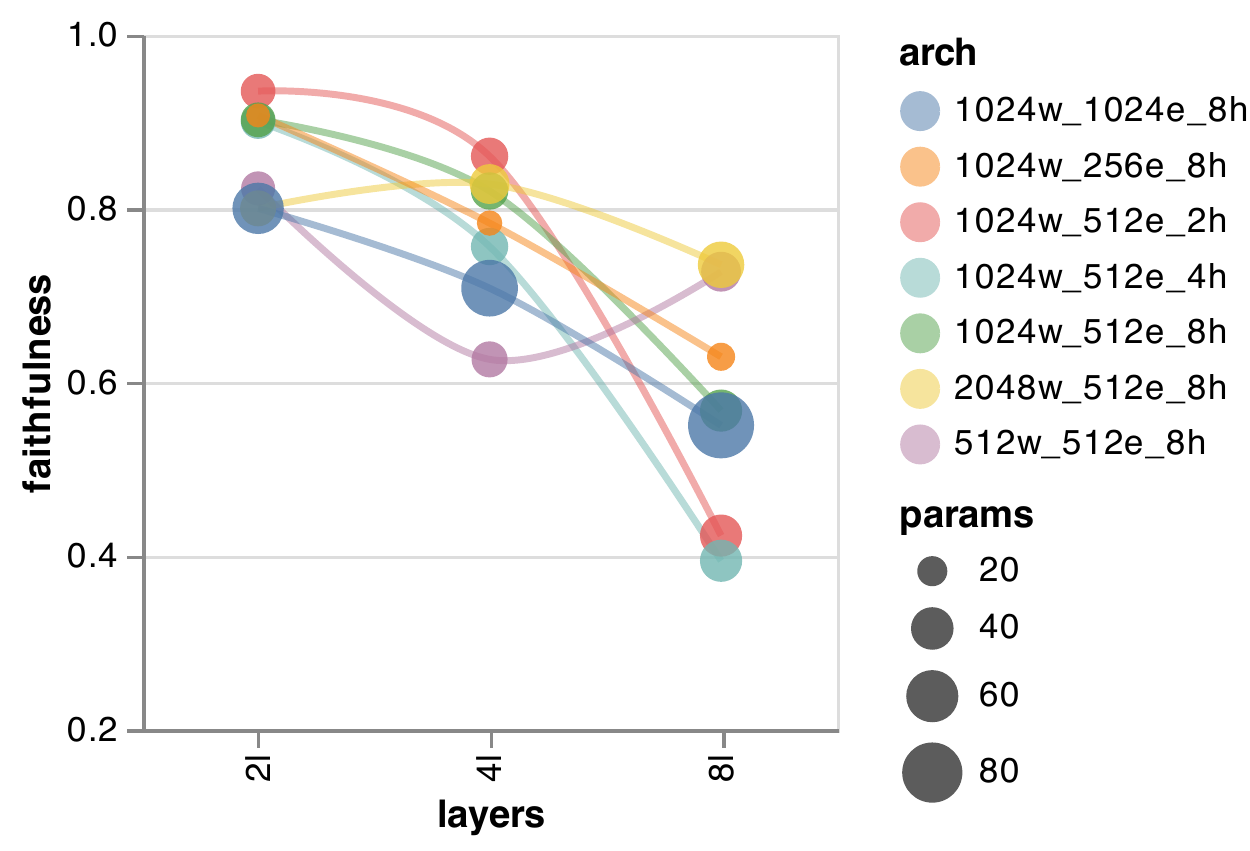} &
\includegraphics[width=0.45\textwidth]{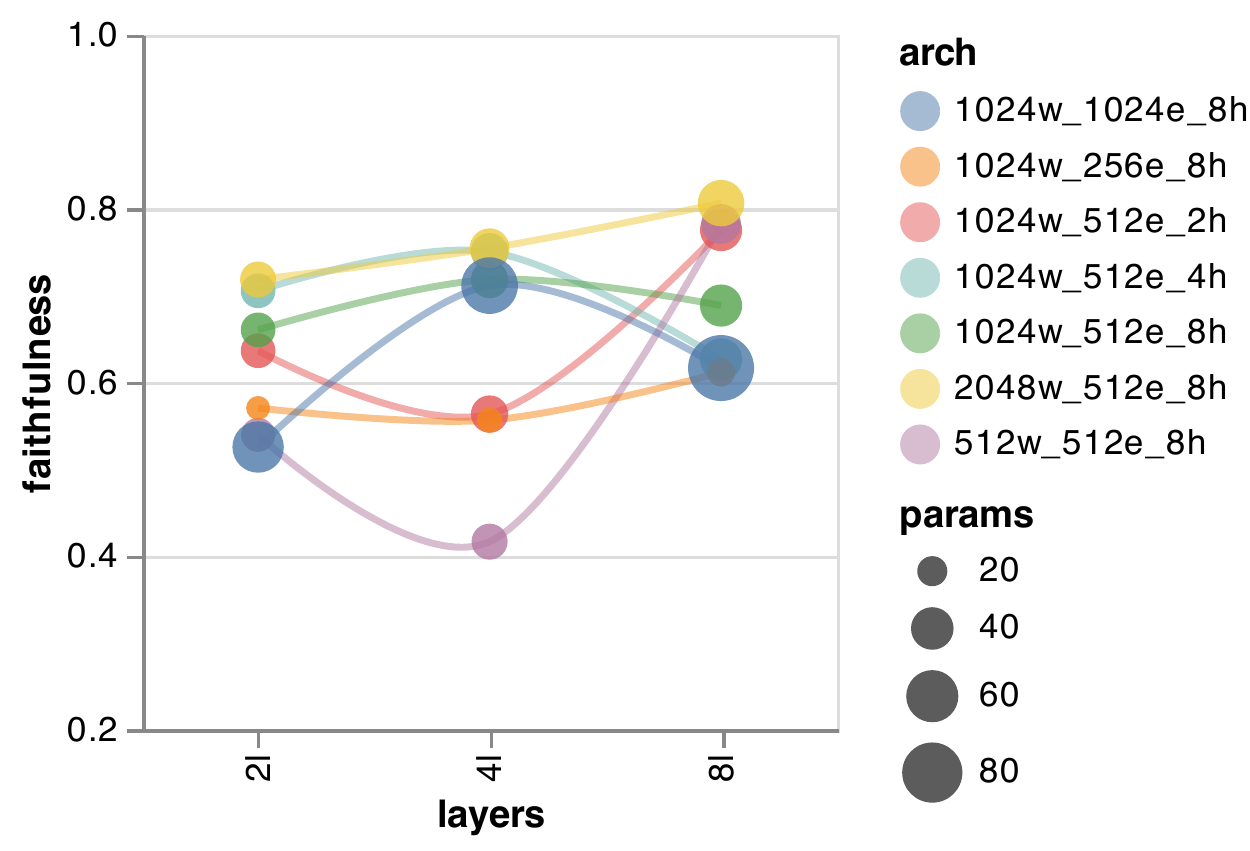} \\
(c) Faithfulness benchmark with SmoothGrad & (d) Faithfulness benchmark with Integrated Gradients \\
\end{tabular}
\caption{Analysis of model configuration vs. plausibility and faithfulness on Syneval benchmark. Each model configuration is color-coded, while the parameter size (in millions) is shown with circle size. \texttt{l}, \texttt{w}, \texttt{e}, \texttt{h} stands for model depth, width of feed-forward layers after self-attention, embedding size, and the number of heads. Note that the faithfulness numbers plotted here are the ones interpreted with expected scenario predictions. \label{fig:config-syneval-all}}
\end{figure*}

As mentioned in Section \ref{sec:analysis}, we would like to know if there are any predictable patterns in how interpretation quality changes with varying model configurations.
To answer this question, we build smaller Transformer language models with various depth, number of heads, embedding size, and feed-forward layer width settings, while keeping other hyperparameters unchanged.

We show two different groups of comparison here.
Figure~\ref{fig:config-sg-all} shows our investigation on the interaction between model configuration and interpretation plausibility on PTB and CoNLL test sets.
In general, Integrated Gradients method works better for deeper models, while SG works better for shallower models on the PTB test set, but remains roughly the same performance for all architectures on the CoNLL test set.
This indicates the noisiness of the trend we are investigating, as both interpretability methods and evaluation dataset choice can influence the trend.
As for the other factors of the model configurations, the trend is even noisier (note how much rankings of different configurations change moving from shallow to deep models) and do not show any clear patterns.

Figure~\ref{fig:config-syneval-all}, on the other hand, focuses on one specific dataset and investigate the trend on both the plausibility and input faithfulness with varying model configurations.
For plausibility results, we largely see the same trend as on PTB dataset.
For faithfulness results, the trend for SG is largely the same as plausibility.
For IG, the variance across other factors of configurations tends to be different on shallower models vs. deeper models, but overall still shows higher numbers for deeper models like for plausibility.

Overall, these analyses further support our conclusion in the main paper, that interpretation qualities are sensitive to model configuration changes, and we reiterate that evaluations of saliency methods should be conducted on the specific model configurations of interest, and trends of interpretation quality on a specific model configuration should not be over-generalized to other configurations.

\section{Language Model Perplexities}

Parameter size and perplexity on WikiText-103 dev set for all language models are shown in Table~\ref{tab:all-ppl} for reference.

Below are the respective commands to reproduce these results.
\begin{itemize}
\item LSTM:
\texttt{python -u main.py --epochs 50 --nlayers 3 --emsize 400 --nhid 2000 --dropoute 0 --dropouth 0.01 --dropouti 0.01 --dropout 0.4 --wdrop 0.2 --bptt 140 --batch\_size 60 --optimizer adam --lr 1e-3 --data data/wikitext-103 --save save --when 25 35 --model LSTM}
\item QRNN:
\texttt{python -u main.py --epochs 14 --nlayers 4 --emsize 400 --nhid 2500 --alpha 0 --beta 0 --dropoute 0 --dropouth 0.1 --dropouti 0.1 --dropout 0.1 --wdrop 0 --wdecay 0 --bptt 140 --batch\_size 40 --optimizer adam --lr 1e-3 --data data/wikitext-103 --save save --when 12 --model QRNN}
\item Transformer:
\texttt{python train.py --task language\_modeling data-bin/wikitext-103 --save-dir checkpoints --arch transformer\_lm\_wiki103 --decoder-layers \$layers --decoder-attention-heads \$num\_heads --decoder-embed-dim \$emb --decoder-ffn-embed-dim \$width --max-update 286000 --max-lr 1.0 --t-mult 2 --lr-period-updates 270000 --lr-scheduler cosine --lr-shrink 0.75 --warmup-updates 16000 --warmup-init-lr 1e-07 --min-lr 1e-09 --optimizer nag --lr 0.0001 --clip-norm 0.1 --criterion adaptive\_loss --max-tokens 3072 --update-freq 3 --tokens-per-sample 3072 --seed 1 --sample-break-mode none --skip-invalid-size-inputs-valid-test --ddp-backend=no\_c10d}
\end{itemize}

\begin{table*}[h]
\centering
\scalebox{0.9}{
\begin{tabular}{@{}lrlrr@{}}
\toprule
\textbf{Architectures} & \multicolumn{1}{c}{\textbf{Layers}} & \multicolumn{1}{c}{\textbf{Config}} & \multicolumn{1}{c}{\textbf{Params (M)}} & \multicolumn{1}{c}{\textbf{dev ppl}} \\ \midrule
\textbf{LSTM}          & 3                                   & -                                   & 162                                     & 37.65                                \\
                       & 2                                   & -                                   & 130                                     & 41.97                                \\
\textbf{QRNN}          & 4                                   & -                                   & 154                                     & 32.12                                \\
                       & 3                                   & -                                   & 135                                     & 36.54                                \\
\textbf{Transformer}   & \multicolumn{1}{l}{\textbf{}}       &                                     & \multicolumn{1}{l}{}                    & \multicolumn{1}{l}{}                 \\
                       & 16                                  & 4096w\_1024e\_8h                    & 247                                     & 17.97                                \\
                       & 4                                   & 4096w\_1024e\_8h (Distill Student)  & 96.1                                    & 24.92                                \\
                       & 4                                   & 4096w\_1024e\_8h                    & 96.1                                    & 28.96                                \\
                       & 8                                   & 1024w\_512e\_2h                     & 39.3                                    & 32.63                                \\
                       & 8                                   & 1024w\_512e\_4h                     & 39.3                                    & 32.09                                \\
                       & 8                                   & 1024w\_512e\_8h                     & 39.3                                    & 31.38                                \\
                       & 8                                   & 512w\_512e\_8h                      & 35.1                                    & 33.99                                \\
                       & 8                                   & 2048w\_512e\_8h                     & 47.7                                    & 30.19                                \\
                       & 8                                   & 1024w\_256e\_8h                     & 17.4                                    & 41.56                                \\
                       & 8                                   & 1024w\_1024e\_8h                    & 96.1                                    & 27.01                                \\
                       & 4                                   & 1024w\_512e\_2h                     & 30.8                                    & 37.03                                \\
                       & 4                                   & 1024w\_512e\_4h                     & 30.8                                    & 35.67                                \\
                       & 4                                   & 1024w\_512e\_8h                     & 30.8                                    & 35.82                                \\
                       & 4                                   & 512w\_512e\_8h                      & 28.7                                    & 38.34                                \\
                       & 4                                   & 2048w\_512e\_8h                     & 35.0                                    & 33.70                                \\
                       & 4                                   & 1024w\_256e\_8h                     & 14.3                                    & 48.47                                \\
                       & 4                                   & 1024w\_1024e\_8h                    & 70.9                                    & 30.46                                \\
                       & 2                                   & 1024w\_512e\_2h                     & 26.6                                    & 44.45                                \\
                       & 2                                   & 1024w\_512e\_4h                     & 26.6                                    & 42.23                                \\
                       & 2                                   & 1024w\_512e\_8h                     & 26.6                                    & 41.86                                \\
                       & 2                                   & 512w\_512e\_8h                      & 25.6                                    & 44.97                                \\
                       & 2                                   & 2048w\_512e\_8h                     & 28.7                                    & 38.99                                \\
                       & 2                                   & 1024w\_256e\_8h                     & 12.7                                    & 59.16                                \\
                       & 2                                   & 1024w\_1024e\_8h                    & 58.3                                    & 36.06                                \\ \bottomrule
\end{tabular}
}
\caption{Parameter size (in millions) and perplexity on WikiText-103 dev set for all language models we trained. \label{tab:all-ppl}}
\end{table*}

\section{Additional Interpretation Examples}

We show some additional interpretations generated by the state-of-the-art LSTM (Table~\ref{tab:add-exp-lstm}), QRNN (Table~\ref{tab:add-exp-qrnn}) and Transformer (Table~\ref{tab:add-exp-transformer}) models on PTB and CoNLL dataset, with their respective best-performing interpretation method.

\setlength{\fboxsep}{2pt}
\begin{table*}[t]
    \centering
    \scalebox{0.8}{
    \begin{tabular}{p{\textwidth}}

\toprule
{\bf PTB} \\
1-\quad{\tt \colorbox{applegreen!1.79736370791!white}{(U.S.)}
\colorbox{applegreen!0.153302543093!white}{(Trade)}
\colorbox{bananamania!1.91812092442!white}{(Representative)}
\colorbox{applegreen!0.682117923509!white}{(Carla)}
\colorbox{applegreen!2.3309945789!white}{(Hills)}
\colorbox{applegreen!0.422603011479!white}{said}
\colorbox{bananamania!0.20009395814!white}{the}
\colorbox{applegreen!3.94461377619!white}{first}
\colorbox{bananamania!3.17456157969!white}{dispute-settlement}
\colorbox{applegreen!0.312517345215!white}{(panel)}
\colorbox{bananamania!0.262344044653!white}{set}
\colorbox{applegreen!3.66860480128!white}{up}
\colorbox{bananamania!0.506038854652!white}{under}
\colorbox{bananamania!1.66140886112!white}{the}
\colorbox{applegreen!0.888961270246!white}{U.S.-Canadian}
\colorbox{bananamania!0.168417312586!white}{``}
\colorbox{bananamania!5.79918318361!white}{free}
\colorbox{bananamania!2.52432541322!white}{(trade)}
\colorbox{applegreen!1.64145107667!white}{''}
\colorbox{applegreen!3.56038302654!white}{(agreement)}
\colorbox{bananamania!25.4864645418!white}{has}
\colorbox{bananamania!3.58979481629!white}{ruled}
\colorbox{applegreen!13.4666813544!white}{that}
\colorbox{bananamania!24.7852437977!white}{(Canada)}
\colorbox{applegreen!17.4602333756!white}{'s}
\colorbox{bananamania!2.46981311212!white}{[restrictions]}
\colorbox{bananamania!23.4448930025!white}{on}
\colorbox{applegreen!10.0470482781!white}{[exports]}
\colorbox{bananamania!0.6775536581!white}{of}
\colorbox{applegreen!13.756533422!white}{(Pacific)}
\colorbox{bananamania!23.1796003791!white}{(salmon)}
\colorbox{bananamania!100.0!white}{and}
\colorbox{bananamania!13.5230685496!white}{(herring)}
| PLURAL}

2-\quad\texttt{\colorbox{bananamania!1.0124066968!white}{Individual}
\colorbox{bananamania!0.766128800886!white}{[investors]}
\colorbox{applegreen!2.00168256281!white}{,}
\colorbox{applegreen!0.897557786196!white}{(investment)}
\colorbox{applegreen!0.259027415993!white}{[firms]}
\colorbox{applegreen!0.526556214494!white}{and}
\colorbox{applegreen!1.33895944539!white}{[arbitragers]}
\colorbox{applegreen!0.998702737116!white}{who}
\colorbox{bananamania!0.861089116362!white}{speculate}
\colorbox{bananamania!3.69114365489!white}{in}
\colorbox{bananamania!3.35920263184!white}{the}
\colorbox{applegreen!0.246922867818!white}{[stocks]}
\colorbox{applegreen!2.40435332073!white}{of}
\colorbox{applegreen!0.963282354802!white}{(takeover)}
\colorbox{bananamania!0.973659307602!white}{[candidates]}
\colorbox{bananamania!0.869777472992!white}{can}
\colorbox{applegreen!0.200134203453!white}{suffer}
\colorbox{applegreen!0.366708078474!white}{(liquidity)}
\colorbox{applegreen!7.12731944731!white}{and}
\colorbox{bananamania!3.48793631411!white}{(payment)}
\colorbox{applegreen!2.84124878143!white}{[problems]}
\colorbox{applegreen!8.78076936769!white}{when}
\colorbox{applegreen!0.807893645992!white}{[stocks]}
\colorbox{bananamania!1.75935443973!white}{dive}
\colorbox{applegreen!50.9735622676!white}{;}
\colorbox{applegreen!20.8918845094!white}{those}
\colorbox{applegreen!66.4800608663!white}{[investors]}
\colorbox{bananamania!100.0!white}{often}
| PLURAL}

3-\quad\texttt{\colorbox{bananamania!22.9789898932!white}{(U.S.)}
\colorbox{applegreen!61.7762209339!white}{[companies]}
\colorbox{bananamania!14.2088528995!white}{wanting}
\colorbox{applegreen!25.7084952491!white}{to}
\colorbox{bananamania!75.3901458236!white}{expand}
\colorbox{applegreen!100.0!white}{in}
\colorbox{applegreen!51.818783671!white}{(Europe)}
| PLURAL
}

4-\quad\texttt{\colorbox{applegreen!0.34484939782!white}{CURBING}
\colorbox{applegreen!1.26468585294!white}{[WAGE]}
\colorbox{applegreen!2.6567754716!white}{(BOOSTS)}
\colorbox{bananamania!3.05425702898!white}{will}
\colorbox{applegreen!2.91872292738!white}{get}
\colorbox{bananamania!1.84714140987!white}{high}
\colorbox{applegreen!0.87872822594!white}{[priority]}
\colorbox{bananamania!2.76033585529!white}{again}
\colorbox{bananamania!6.84316105069!white}{in}
\colorbox{bananamania!0.206044185334!white}{1990}
\colorbox{applegreen!2.09034294034!white}{collective}
\colorbox{bananamania!0.163746225015!white}{[bargaining]}
\colorbox{bananamania!2.17300643099!white}{,}
\colorbox{bananamania!29.0872709192!white}{a}
\colorbox{applegreen!2.34801789188!white}{[Bureau]}
\colorbox{applegreen!15.815353176!white}{of}
\colorbox{bananamania!3.53690050875!white}{[National]}
\colorbox{bananamania!0.503861370695!white}{[Affairs]}
\colorbox{applegreen!7.34133044684!white}{[survey]}
\colorbox{bananamania!18.8961668226!white}{of}
\colorbox{applegreen!5.22196686667!white}{250}
\colorbox{applegreen!23.8689816362!white}{(companies)}
\colorbox{bananamania!5.69090008695!white}{with}
\colorbox{bananamania!0.748091419543!white}{(pacts)}
\colorbox{bananamania!1.07120293841!white}{expiring}
\colorbox{applegreen!100.0!white}{next}
\colorbox{bananamania!51.6176589243!white}{[year]}
| PLURAL
}

5-\quad\texttt{
\colorbox{bananamania!11.7957320442!white}{TEMPORARY}
\colorbox{bananamania!11.7614502891!white}{(WORKERS)}
\colorbox{bananamania!16.4660584796!white}{have}
\colorbox{applegreen!16.9511014375!white}{good}
\colorbox{applegreen!1.3395357377!white}{(educations)}
\colorbox{applegreen!90.1974490485!white}{,}
\colorbox{applegreen!100.0!white}{the}
\colorbox{applegreen!81.3703358749!white}{[National]}
\colorbox{bananamania!26.2150394761!white}{[Association]}
\colorbox{applegreen!22.2280227463!white}{of}
\colorbox{applegreen!68.6449753269!white}{[Temporary]}
\colorbox{applegreen!12.913472234!white}{[Services]}
| SINGULAR
}
\\\midrule

{\bf CoNLL} \\
1-\quad\texttt{
\colorbox{applegreen!1.66761011418!white}{[Israeli]}
\colorbox{applegreen!0.0435974262967!white}{[Prime]}
\colorbox{applegreen!0.283112246002!white}{[Minister]}
\colorbox{applegreen!0.0612991523103!white}{[Ehud]}
\colorbox{bananamania!0.0339742225931!white}{[Barak]}
\colorbox{bananamania!0.32046769015!white}{says}
\colorbox{applegreen!5.00378634586!white}{[he]}
\colorbox{applegreen!1.29675379381!white}{is}
\colorbox{bananamania!0.488928064925!white}{freezing}
\colorbox{bananamania!0.575723226147!white}{tens}
\colorbox{applegreen!0.202213356393!white}{of}
\colorbox{applegreen!0.341078448454!white}{millions}
\colorbox{bananamania!1.84677111091!white}{of}
\colorbox{bananamania!2.90605706741!white}{dollars}
\colorbox{applegreen!8.74973060091!white}{in}
\colorbox{bananamania!0.298816259456!white}{tax}
\colorbox{applegreen!0.699252339873!white}{payments}
\colorbox{bananamania!0.300952226113!white}{to}
\colorbox{applegreen!5.38785792522!white}{the}
\colorbox{applegreen!0.441899454576!white}{Palestinian}
\colorbox{bananamania!2.91068233922!white}{Authority}
\colorbox{bananamania!3.06690627766!white}{.}
\colorbox{applegreen!1.59189806573!white}{[Mr.]}
\colorbox{bananamania!0.357336149457!white}{[Barak]}
\colorbox{applegreen!4.83242109911!white}{says}
\colorbox{applegreen!10.5809852438!white}{[he]}
\colorbox{bananamania!1.63204084328!white}{is}
\colorbox{applegreen!1.09812393546!white}{withholding}
\colorbox{bananamania!3.17276932706!white}{the}
\colorbox{applegreen!11.9827129929!white}{money}
\colorbox{bananamania!0.773462992275!white}{until}
\colorbox{applegreen!0.611535524819!white}{the}
\colorbox{applegreen!0.0379021921032!white}{Palestinians}
\colorbox{applegreen!0.145372288894!white}{abide}
\colorbox{applegreen!3.61592422436!white}{by}
\colorbox{bananamania!0.62202574757!white}{cease}
\colorbox{bananamania!3.8946168136!white}{-}
\colorbox{applegreen!11.1758528012!white}{fire}
\colorbox{applegreen!5.52185643861!white}{agreements}
\colorbox{applegreen!3.73572944771!white}{.}
\colorbox{applegreen!3.9016007042!white}{Earlier}
\colorbox{applegreen!12.5395794185!white}{Thursday}
\colorbox{applegreen!7.61001990161!white}{[Mr.]}
\colorbox{bananamania!0.0746830266644!white}{[Barak]}
\colorbox{applegreen!6.22174613698!white}{ruled}
\colorbox{bananamania!7.09581952966!white}{out}
\colorbox{bananamania!2.70067330168!white}{an}
\colorbox{applegreen!4.20095144578!white}{early}
\colorbox{applegreen!0.833959457684!white}{resumption}
\colorbox{bananamania!4.58238107473!white}{of}
\colorbox{applegreen!6.37384318092!white}{peace}
\colorbox{bananamania!1.13545033054!white}{talks}
\colorbox{bananamania!26.7021749208!white}{,}
\colorbox{bananamania!7.35203557777!white}{even}
\colorbox{applegreen!18.2829169344!white}{with}
\colorbox{applegreen!3.99015601842!white}{the}
\colorbox{bananamania!1.26735916992!white}{United}
\colorbox{applegreen!17.3342565594!white}{States}
\colorbox{applegreen!11.4856188024!white}{acting}
\colorbox{applegreen!14.8267512725!white}{as}
\colorbox{bananamania!1.69183028752!white}{intermediary}
\colorbox{bananamania!18.914668818!white}{.}
\colorbox{bananamania!21.4346120605!white}{(Eve)}
\colorbox{applegreen!13.9350028323!white}{(Conette)}
\colorbox{applegreen!4.14687117516!white}{reports}
\colorbox{applegreen!9.61659266672!white}{from}
\colorbox{applegreen!4.98519329286!white}{Jerusalem}
\colorbox{applegreen!100.0!white}{.}
\colorbox{applegreen!16.1418682049!white}{Defending}
\colorbox{applegreen!53.6297199139!white}{what}
| MALE
}

2-\quad\texttt{\colorbox{bananamania!3.68962746996!white}{Once}
\colorbox{bananamania!2.25844701432!white}{again}
\colorbox{applegreen!2.51024305723!white}{there}
\colorbox{applegreen!0.646120930539!white}{'ll}
\colorbox{bananamania!3.96807170613!white}{be}
\colorbox{applegreen!2.68175026876!white}{two}
\colorbox{applegreen!0.275988394421!white}{presidential}
\colorbox{applegreen!0.263189664988!white}{candidates}
\colorbox{bananamania!0.198086519886!white}{missing}
\colorbox{applegreen!1.14052786019!white}{from}
\colorbox{bananamania!4.38025404914!white}{the}
\colorbox{applegreen!0.0704641442202!white}{debate}
\colorbox{applegreen!4.51421293972!white}{.}
\colorbox{bananamania!1.53647895323!white}{Pat}
\colorbox{bananamania!2.71083286662!white}{Buchanan}
\colorbox{bananamania!0.0528162156983!white}{hardly}
\colorbox{applegreen!1.25312806281!white}{registers}
\colorbox{bananamania!5.23588201237!white}{on}
\colorbox{applegreen!3.65637261292!white}{the}
\colorbox{bananamania!6.21251555042!white}{political}
\colorbox{applegreen!1.78946177152!white}{radar}
\colorbox{applegreen!1.75411148079!white}{this}
\colorbox{bananamania!2.67880740779!white}{year}
\colorbox{applegreen!2.68900520185!white}{.}
\colorbox{bananamania!3.44491762906!white}{And}
\colorbox{bananamania!3.54499200149!white}{Ralph}
\colorbox{applegreen!0.731599185303!white}{Nader}
\colorbox{applegreen!1.23984042819!white}{,}
\colorbox{applegreen!9.13827860713!white}{who}
\colorbox{applegreen!2.70447012107!white}{may}
\colorbox{applegreen!14.2411869574!white}{make}
\colorbox{bananamania!2.09502971578!white}{the}
\colorbox{applegreen!3.86073435383!white}{difference}
\colorbox{applegreen!0.261178203924!white}{between}
\colorbox{applegreen!6.65275477738!white}{a}
\colorbox{applegreen!2.04708834045!white}{[Gore]}
\colorbox{applegreen!0.130616802084!white}{or}
\colorbox{bananamania!2.72688570424!white}{[Bush]}
\colorbox{bananamania!8.10570028373!white}{win}
\colorbox{bananamania!5.41415456028!white}{in}
\colorbox{applegreen!2.4668145465!white}{several}
\colorbox{bananamania!5.64275624189!white}{places}
\colorbox{applegreen!1.97708317743!white}{.}
\colorbox{bananamania!11.5156868094!white}{(ABC)}
\colorbox{bananamania!15.6383335265!white}{('s)}
\colorbox{bananamania!3.47838371427!white}{(Linda)}
\colorbox{bananamania!39.1782661105!white}{(Douglas)}
\colorbox{applegreen!81.1338969303!white}{was}
\colorbox{bananamania!100.0!white}{with}
| FEMALE
} \\

\bottomrule

    \end{tabular}
    }
    \caption{Addition interpretation examples with LSTM.}
    \label{tab:add-exp-lstm}
\end{table*}

\begin{table*}[t]
    \centering
    \scalebox{0.8}{
    \begin{tabular}{p{\textwidth}}

\toprule
{\bf PTB} \\
1-\quad{\tt \colorbox{applegreen!36.3288860558!white}{(U.S.)}
\colorbox{bananamania!18.6323538329!white}{(Trade)}
\colorbox{bananamania!15.0393693388!white}{(Representative)}
\colorbox{bananamania!14.7149284423!white}{(Carla)}
\colorbox{bananamania!2.80847614163!white}{(Hills)}
\colorbox{bananamania!20.303809465!white}{said}
\colorbox{applegreen!0.689179052792!white}{the}
\colorbox{bananamania!13.7054425621!white}{first}
\colorbox{bananamania!7.323037989!white}{dispute-settlement}
\colorbox{bananamania!7.90348457053!white}{(panel)}
\colorbox{applegreen!1.03932867096!white}{set}
\colorbox{bananamania!4.68251976171!white}{up}
\colorbox{bananamania!14.6931949274!white}{under}
\colorbox{bananamania!13.5053803417!white}{the}
\colorbox{bananamania!34.2681162771!white}{U.S.-Canadian}
\colorbox{bananamania!4.97928655349!white}{``}
\colorbox{applegreen!27.2576250855!white}{free}
\colorbox{applegreen!22.1039282711!white}{(trade)}
\colorbox{applegreen!9.64694455299!white}{''}
\colorbox{applegreen!16.5920581545!white}{(agreement)}
\colorbox{applegreen!14.4086284753!white}{has}
\colorbox{bananamania!5.75814835506!white}{ruled}
\colorbox{bananamania!1.43031840834!white}{that}
\colorbox{bananamania!1.15278113579!white}{(Canada)}
\colorbox{bananamania!1.15277946104!white}{'s}
\colorbox{applegreen!14.3672725918!white}{[restrictions]}
\colorbox{applegreen!23.9913658516!white}{on}
\colorbox{applegreen!33.0289326963!white}{[exports]}
\colorbox{applegreen!11.6539384766!white}{of}
\colorbox{bananamania!1.50261919!white}{(Pacific)}
\colorbox{applegreen!22.3663477834!white}{(salmon)}
\colorbox{applegreen!100.0!white}{and}
\colorbox{bananamania!93.6573966606!white}{(herring)}
| PLURAL}

2-\quad\texttt{\colorbox{applegreen!6.89347527347!white}{Individual}
\colorbox{bananamania!0.819512284672!white}{[investors]}
\colorbox{bananamania!3.57832972823!white}{,}
\colorbox{bananamania!1.25957756836!white}{(investment)}
\colorbox{applegreen!0.874959065685!white}{[firms]}
\colorbox{applegreen!0.12952276219!white}{and}
\colorbox{bananamania!1.27686336439!white}{[arbitragers]}
\colorbox{bananamania!0.737915873443!white}{who}
\colorbox{bananamania!1.38492590212!white}{speculate}
\colorbox{bananamania!0.832221556253!white}{in}
\colorbox{bananamania!2.04744059661!white}{the}
\colorbox{bananamania!1.37463820732!white}{[stocks]}
\colorbox{applegreen!2.01028753976!white}{of}
\colorbox{bananamania!3.22807880198!white}{(takeover)}
\colorbox{applegreen!0.271950164962!white}{[candidates]}
\colorbox{bananamania!2.50048692425!white}{can}
\colorbox{bananamania!3.07864961757!white}{suffer}
\colorbox{bananamania!2.10096827411!white}{(liquidity)}
\colorbox{bananamania!1.00860912847!white}{and}
\colorbox{bananamania!0.335214780794!white}{(payment)}
\colorbox{applegreen!0.123039228327!white}{[problems]}
\colorbox{bananamania!9.48021571099!white}{when}
\colorbox{applegreen!0.696309357471!white}{[stocks]}
\colorbox{applegreen!9.17416335018!white}{dive}
\colorbox{bananamania!27.8906871961!white}{;}
\colorbox{bananamania!7.03762166815!white}{those}
\colorbox{applegreen!100.0!white}{[investors]}
\colorbox{bananamania!20.7188567888!white}{often}
| PLURAL}

3-\quad\texttt{\colorbox{applegreen!68.3200244587!white}{(U.S.)}
\colorbox{bananamania!12.7641064652!white}{[companies]}
\colorbox{applegreen!0.54389578051!white}{wanting}
\colorbox{applegreen!39.9084164248!white}{to}
\colorbox{applegreen!63.309503942!white}{expand}
\colorbox{applegreen!100.0!white}{in}
\colorbox{bananamania!58.606656306!white}{(Europe)}
| PLURAL
}

4-\quad\texttt{\colorbox{applegreen!0.800254900324!white}{CURBING}
\colorbox{bananamania!1.47737216321!white}{[WAGE]}
\colorbox{bananamania!2.80650978053!white}{(BOOSTS)}
\colorbox{applegreen!7.80205619174!white}{will}
\colorbox{applegreen!5.07727105172!white}{get}
\colorbox{applegreen!7.99428362978!white}{high}
\colorbox{bananamania!19.3176687086!white}{[priority]}
\colorbox{applegreen!16.2130164018!white}{again}
\colorbox{bananamania!0.145440619536!white}{in}
\colorbox{applegreen!6.24336074603!white}{1990}
\colorbox{applegreen!7.96665168318!white}{collective}
\colorbox{bananamania!10.5397564771!white}{[bargaining]}
\colorbox{applegreen!0.180032834812!white}{,}
\colorbox{bananamania!2.45061594514!white}{a}
\colorbox{bananamania!25.091909855!white}{[Bureau]}
\colorbox{applegreen!20.9587667713!white}{of}
\colorbox{bananamania!0.508625198739!white}{[National]}
\colorbox{bananamania!7.48811251143!white}{[Affairs]}
\colorbox{applegreen!60.1562962396!white}{[survey]}
\colorbox{bananamania!18.5264131783!white}{of}
\colorbox{applegreen!6.4224993457!white}{250}
\colorbox{applegreen!18.0304047435!white}{(companies)}
\colorbox{applegreen!8.80636922022!white}{with}
\colorbox{applegreen!31.638433021!white}{(pacts)}
\colorbox{bananamania!100.0!white}{expiring}
\colorbox{applegreen!5.92638767823!white}{next}
\colorbox{bananamania!19.1208039931!white}{[year]}
| PLURAL
}

5-\quad\texttt{
\colorbox{applegreen!4.40441157234!white}{TEMPORARY}
\colorbox{applegreen!18.4207001415!white}{(WORKERS)}
\colorbox{applegreen!1.71829586762!white}{have}
\colorbox{applegreen!15.3559033394!white}{good}
\colorbox{bananamania!100.0!white}{(educations)}
\colorbox{applegreen!15.6140356868!white}{,}
\colorbox{bananamania!4.90692147104!white}{the}
\colorbox{applegreen!1.93746188265!white}{[National]}
\colorbox{applegreen!45.8938421383!white}{[Association]}
\colorbox{bananamania!24.4422128881!white}{of}
\colorbox{applegreen!95.4130260866!white}{[Temporary]}
\colorbox{applegreen!21.3908533964!white}{[Services]}
| PLURAL
}
\\\midrule

{\bf CoNLL} \\
1-\quad\texttt{
\colorbox{applegreen!9.31993163362!white}{[Israeli]}
\colorbox{bananamania!0.71604312377!white}{[Prime]}
\colorbox{bananamania!10.174707222!white}{[Minister]}
\colorbox{applegreen!7.79396314067!white}{[Ehud]}
\colorbox{bananamania!5.22212200043!white}{[Barak]}
\colorbox{applegreen!4.45624674944!white}{says}
\colorbox{bananamania!10.8184804856!white}{[he]}
\colorbox{applegreen!0.0499781484063!white}{is}
\colorbox{applegreen!0.843251812586!white}{freezing}
\colorbox{applegreen!0.88110063968!white}{tens}
\colorbox{bananamania!0.105212784185!white}{of}
\colorbox{applegreen!1.63630889394!white}{millions}
\colorbox{applegreen!0.575661912275!white}{of}
\colorbox{applegreen!1.1658030772!white}{dollars}
\colorbox{applegreen!0.302628567488!white}{in}
\colorbox{bananamania!2.49902567963!white}{tax}
\colorbox{applegreen!1.62446123144!white}{payments}
\colorbox{applegreen!0.392264427627!white}{to}
\colorbox{bananamania!3.26455544016!white}{the}
\colorbox{applegreen!1.28643109624!white}{Palestinian}
\colorbox{applegreen!5.68197234244!white}{Authority}
\colorbox{bananamania!5.3636920956!white}{.}
\colorbox{bananamania!16.0143567332!white}{[Mr.]}
\colorbox{bananamania!5.85075129383!white}{[Barak]}
\colorbox{applegreen!7.19526867236!white}{says}
\colorbox{bananamania!16.5316998052!white}{[he]}
\colorbox{applegreen!0.721014950048!white}{is}
\colorbox{applegreen!1.51629323514!white}{withholding}
\colorbox{bananamania!2.72878310392!white}{the}
\colorbox{applegreen!0.680594387885!white}{money}
\colorbox{applegreen!1.19889230303!white}{until}
\colorbox{bananamania!3.6541016085!white}{the}
\colorbox{bananamania!1.33525547235!white}{Palestinians}
\colorbox{applegreen!2.31935742475!white}{abide}
\colorbox{applegreen!1.09317950867!white}{by}
\colorbox{applegreen!1.32096751873!white}{cease}
\colorbox{applegreen!1.05338058364!white}{-}
\colorbox{applegreen!3.50346675191!white}{fire}
\colorbox{applegreen!3.02373351026!white}{agreements}
\colorbox{applegreen!11.6750616386!white}{.}
\colorbox{bananamania!7.32445939876!white}{Earlier}
\colorbox{applegreen!9.90603933724!white}{Thursday}
\colorbox{bananamania!23.4159965654!white}{[Mr.]}
\colorbox{bananamania!18.1909107008!white}{[Barak]}
\colorbox{bananamania!4.56699618131!white}{ruled}
\colorbox{applegreen!4.38208982343!white}{out}
\colorbox{bananamania!0.859049232489!white}{an}
\colorbox{applegreen!2.68136042953!white}{early}
\colorbox{bananamania!3.85152864905!white}{resumption}
\colorbox{applegreen!2.48849677742!white}{of}
\colorbox{applegreen!0.654967857628!white}{peace}
\colorbox{applegreen!3.72934702432!white}{talks}
\colorbox{bananamania!0.883375640975!white}{,}
\colorbox{bananamania!3.48032549939!white}{even}
\colorbox{bananamania!6.17160907909!white}{with}
\colorbox{bananamania!7.24451436384!white}{the}
\colorbox{bananamania!8.71690828339!white}{United}
\colorbox{bananamania!6.41633357704!white}{States}
\colorbox{bananamania!2.11332023332!white}{acting}
\colorbox{bananamania!5.10868895999!white}{as}
\colorbox{bananamania!26.0438061433!white}{intermediary}
\colorbox{applegreen!32.7909030339!white}{.}
\colorbox{applegreen!100.0!white}{(Eve)}
\colorbox{applegreen!24.9247301847!white}{(Conette)}
\colorbox{applegreen!11.5171750407!white}{reports}
\colorbox{bananamania!11.8258018876!white}{from}
\colorbox{applegreen!4.31700206431!white}{Jerusalem}
\colorbox{bananamania!22.8090638695!white}{.}
\colorbox{applegreen!41.4691028306!white}{Defending}
\colorbox{bananamania!11.3603776103!white}{what}
| FEMALE
}

2-\quad\texttt{\colorbox{bananamania!11.7292718246!white}{Once}
\colorbox{bananamania!4.66900636526!white}{again}
\colorbox{bananamania!8.87663277669!white}{there}
\colorbox{applegreen!0.505677890655!white}{'ll}
\colorbox{bananamania!5.5563291842!white}{be}
\colorbox{bananamania!2.91608107327!white}{two}
\colorbox{bananamania!10.0529585167!white}{presidential}
\colorbox{bananamania!4.06247292117!white}{candidates}
\colorbox{applegreen!6.39521817249!white}{missing}
\colorbox{bananamania!2.09096501363!white}{from}
\colorbox{bananamania!3.20573300923!white}{the}
\colorbox{bananamania!5.42553734046!white}{debate}
\colorbox{applegreen!20.8444038606!white}{.}
\colorbox{applegreen!2.40018058181!white}{Pat}
\colorbox{applegreen!4.33348842815!white}{Buchanan}
\colorbox{bananamania!2.32536407761!white}{hardly}
\colorbox{applegreen!3.14354570942!white}{registers}
\colorbox{applegreen!3.60908917035!white}{on}
\colorbox{bananamania!2.01858025959!white}{the}
\colorbox{bananamania!15.6586446237!white}{political}
\colorbox{applegreen!5.60817363182!white}{radar}
\colorbox{bananamania!2.11535226795!white}{this}
\colorbox{bananamania!5.55322882264!white}{year}
\colorbox{bananamania!7.64537538318!white}{.}
\colorbox{bananamania!5.07792739032!white}{And}
\colorbox{bananamania!7.9964030763!white}{Ralph}
\colorbox{bananamania!1.42777597033!white}{Nader}
\colorbox{bananamania!0.399642620293!white}{,}
\colorbox{bananamania!11.346349765!white}{who}
\colorbox{bananamania!5.01434037813!white}{may}
\colorbox{applegreen!2.56750719354!white}{make}
\colorbox{applegreen!2.22977649397!white}{the}
\colorbox{bananamania!0.569006173951!white}{difference}
\colorbox{applegreen!0.936271161064!white}{between}
\colorbox{bananamania!0.216617413345!white}{a}
\colorbox{bananamania!5.02877054395!white}{[Gore]}
\colorbox{applegreen!5.31094400879!white}{or}
\colorbox{bananamania!3.39218610001!white}{[Bush]}
\colorbox{bananamania!11.3602027242!white}{win}
\colorbox{applegreen!2.08820341152!white}{in}
\colorbox{bananamania!6.16107509663!white}{several}
\colorbox{applegreen!3.65665362805!white}{places}
\colorbox{applegreen!89.6819090439!white}{.}
\colorbox{applegreen!2.62800891843!white}{(ABC)}
\colorbox{bananamania!12.7702127348!white}{('s)}
\colorbox{applegreen!100.0!white}{(Linda)}
\colorbox{bananamania!4.09705625933!white}{(Douglas)}
\colorbox{applegreen!54.3784823156!white}{was}
\colorbox{bananamania!37.7209676356!white}{with}
| FEMALE }
\\
\bottomrule

    \end{tabular}
    }
    \caption{Addition interpretation examples with QRNN.}
    \label{tab:add-exp-qrnn}
\end{table*}

\setlength{\fboxsep}{2pt}
\begin{table*}[t]
    \centering
    \scalebox{0.8}{
    \begin{tabular}{p{\textwidth}}

\toprule
{\bf PTB} \\
1-\quad{\tt \colorbox{bananamania!6.54997376508!white}{(U.S.)}
\colorbox{applegreen!0.296104667871!white}{(Trade)}
\colorbox{bananamania!3.37567098617!white}{(Representative)}
\colorbox{bananamania!2.49958171114!white}{(Carla)}
\colorbox{bananamania!1.2083287119!white}{(Hills)}
\colorbox{bananamania!6.81395883273!white}{said}
\colorbox{applegreen!4.55599304745!white}{the}
\colorbox{bananamania!2.59409813047!white}{first}
\colorbox{applegreen!0.929035801021!white}{dispute-settlement}
\colorbox{bananamania!1.06642800796!white}{(panel)}
\colorbox{bananamania!2.16652062421!white}{set}
\colorbox{applegreen!1.10345645019!white}{up}
\colorbox{applegreen!5.38532369366!white}{under}
\colorbox{applegreen!12.5556158073!white}{the}
\colorbox{bananamania!0.879667821768!white}{U.S.-Canadian}
\colorbox{applegreen!1.15277173436!white}{``}
\colorbox{applegreen!0.715553937769!white}{free}
\colorbox{applegreen!0.254596003088!white}{(trade)}
\colorbox{bananamania!0.9097812533!white}{''}
\colorbox{applegreen!0.384147437145!white}{(agreement)}
\colorbox{applegreen!17.0889156775!white}{has}
\colorbox{bananamania!1.63536503178!white}{ruled}
\colorbox{applegreen!62.2551135184!white}{that}
\colorbox{bananamania!1.06910098575!white}{(Canada)}
\colorbox{applegreen!12.1228340716!white}{'s}
\colorbox{applegreen!4.30509366348!white}{[restrictions]}
\colorbox{bananamania!2.6887397596!white}{on}
\colorbox{applegreen!14.442083453!white}{[exports]}
\colorbox{applegreen!21.6900993997!white}{of}
\colorbox{bananamania!1.68510164067!white}{(Pacific)}
\colorbox{applegreen!8.60557784416!white}{(salmon)}
\colorbox{applegreen!100.0!white}{and}
\colorbox{bananamania!34.352128446!white}{(herring)}
| PLURAL}

2-\quad\texttt{\colorbox{applegreen!1.2487016437!white}{Individual}
\colorbox{bananamania!1.14759023921!white}{[investors]}
\colorbox{bananamania!1.15852974083!white}{,}
\colorbox{applegreen!0.820023175718!white}{(investment)}
\colorbox{bananamania!1.43586119503!white}{[firms]}
\colorbox{bananamania!1.00860516542!white}{and}
\colorbox{bananamania!0.153884250571!white}{[arbitragers]}
\colorbox{applegreen!2.69085565236!white}{who}
\colorbox{applegreen!1.67036334604!white}{speculate}
\colorbox{applegreen!0.131235249606!white}{in}
\colorbox{applegreen!3.89781359441!white}{the}
\colorbox{applegreen!0.805275673463!white}{[stocks]}
\colorbox{applegreen!0.23115875854!white}{of}
\colorbox{applegreen!2.53141793149!white}{(takeover)}
\colorbox{applegreen!0.279448018055!white}{[candidates]}
\colorbox{bananamania!0.273675151403!white}{can}
\colorbox{applegreen!2.66984789146!white}{suffer}
\colorbox{bananamania!1.11715832271!white}{(liquidity)}
\colorbox{applegreen!0.788279265265!white}{and}
\colorbox{applegreen!0.379588486387!white}{(payment)}
\colorbox{applegreen!1.96524828116!white}{[problems]}
\colorbox{applegreen!3.18026661404!white}{when}
\colorbox{applegreen!9.72386893865!white}{[stocks]}
\colorbox{applegreen!5.60821005945!white}{dive}
\colorbox{applegreen!7.18185319297!white}{;}
\colorbox{applegreen!21.2791840438!white}{those}
\colorbox{applegreen!28.79784945!white}{[investors]}
\colorbox{applegreen!100.0!white}{often}
| PLURAL}

3-\quad\texttt{\colorbox{bananamania!85.2270373142!white}{(U.S.)}
\colorbox{applegreen!71.1828633956!white}{[companies]}
\colorbox{applegreen!3.5868207923!white}{wanting}
\colorbox{applegreen!3.99166149881!white}{to}
\colorbox{applegreen!6.02394457813!white}{expand}
\colorbox{applegreen!12.6632794753!white}{in}
\colorbox{bananamania!100.0!white}{(Europe)}
| PLURAL
}

4-\quad\texttt{\colorbox{bananamania!1.2119165992!white}{CURBING}
\colorbox{applegreen!1.28401820417!white}{[WAGE]}
\colorbox{applegreen!1.68564211424!white}{(BOOSTS)}
\colorbox{applegreen!5.26942078392!white}{will}
\colorbox{applegreen!6.97379811225!white}{get}
\colorbox{applegreen!4.34707263864!white}{high}
\colorbox{bananamania!0.646990275564!white}{[priority]}
\colorbox{bananamania!2.79844800909!white}{again}
\colorbox{bananamania!0.279944615411!white}{in}
\colorbox{applegreen!1.99876959568!white}{1990}
\colorbox{applegreen!0.0735327470108!white}{collective}
\colorbox{applegreen!1.78002972041!white}{[bargaining]}
\colorbox{applegreen!2.19956583285!white}{,}
\colorbox{bananamania!0.15490496165!white}{a}
\colorbox{applegreen!0.137531670535!white}{[Bureau]}
\colorbox{applegreen!0.448097351208!white}{of}
\colorbox{applegreen!0.67863965206!white}{[National]}
\colorbox{bananamania!1.37551299733!white}{[Affairs]}
\colorbox{bananamania!1.7940055643!white}{[survey]}
\colorbox{bananamania!11.6971942146!white}{of}
\colorbox{applegreen!3.80384743017!white}{250}
\colorbox{applegreen!6.40702536239!white}{(companies)}
\colorbox{applegreen!8.1301108217!white}{with}
\colorbox{applegreen!4.79437170355!white}{(pacts)}
\colorbox{applegreen!2.3013245142!white}{expiring}
\colorbox{applegreen!34.7149283795!white}{next}
\colorbox{applegreen!100.0!white}{[year]}
| PLURAL
}

5-\quad\texttt{
\colorbox{bananamania!13.7359655229!white}{TEMPORARY}
\colorbox{bananamania!8.22459143804!white}{(WORKERS)}
\colorbox{applegreen!16.3176787204!white}{have}
\colorbox{applegreen!14.5450172939!white}{good}
\colorbox{applegreen!11.722921699!white}{(educations)}
\colorbox{applegreen!100.0!white}{,}
\colorbox{applegreen!43.3176560825!white}{the}
\colorbox{bananamania!17.6400454348!white}{[National]}
\colorbox{applegreen!12.3971165324!white}{[Association]}
\colorbox{applegreen!27.6154730298!white}{of}
\colorbox{bananamania!16.0596955588!white}{[Temporary]}
\colorbox{bananamania!4.19547689863!white}{[Services]}
| SINGULAR
}
\\\midrule

{\bf CoNLL} \\
1-\quad\texttt{
\colorbox{bananamania!5.89122852426!white}{[Israeli]}
\colorbox{bananamania!2.09855245439!white}{[Prime]}
\colorbox{bananamania!8.48593105512!white}{[Minister]}
\colorbox{bananamania!5.7592133362!white}{[Ehud]}
\colorbox{applegreen!2.69830629874!white}{[Barak]}
\colorbox{bananamania!1.74772231961!white}{says}
\colorbox{bananamania!32.3477558554!white}{[he]}
\colorbox{applegreen!3.09946245097!white}{is}
\colorbox{bananamania!1.66832464081!white}{freezing}
\colorbox{bananamania!4.30292457007!white}{tens}
\colorbox{bananamania!3.49778388931!white}{of}
\colorbox{applegreen!3.41055577739!white}{millions}
\colorbox{bananamania!2.71494587009!white}{of}
\colorbox{applegreen!2.30523442278!white}{dollars}
\colorbox{applegreen!5.79687492304!white}{in}
\colorbox{bananamania!6.51085116888!white}{tax}
\colorbox{bananamania!0.188281140903!white}{payments}
\colorbox{bananamania!1.5643893943!white}{to}
\colorbox{applegreen!7.9863387614!white}{the}
\colorbox{applegreen!9.53137643494!white}{Palestinian}
\colorbox{bananamania!0.723105489002!white}{Authority}
\colorbox{bananamania!5.61996016661!white}{.}
\colorbox{bananamania!12.9041635655!white}{[Mr.]}
\colorbox{bananamania!0.795436867074!white}{[Barak]}
\colorbox{bananamania!3.71260943571!white}{says}
\colorbox{bananamania!66.2315192227!white}{[he]}
\colorbox{applegreen!3.49216131443!white}{is}
\colorbox{bananamania!1.79574518861!white}{withholding}
\colorbox{bananamania!1.05520029033!white}{the}
\colorbox{applegreen!2.93310974252!white}{money}
\colorbox{applegreen!5.87097177288!white}{until}
\colorbox{bananamania!7.83792143531!white}{the}
\colorbox{applegreen!0.479245231669!white}{Palestinians}
\colorbox{bananamania!1.57766110192!white}{abide}
\colorbox{bananamania!4.3570955813!white}{by}
\colorbox{applegreen!0.0905770972623!white}{cease}
\colorbox{bananamania!5.58220577933!white}{-}
\colorbox{applegreen!5.43496633979!white}{fire}
\colorbox{bananamania!1.52187094937!white}{agreements}
\colorbox{bananamania!22.8404894272!white}{.}
\colorbox{applegreen!0.78935436417!white}{Earlier}
\colorbox{bananamania!1.9234205981!white}{Thursday}
\colorbox{bananamania!15.324159085!white}{[Mr.]}
\colorbox{applegreen!1.59181531909!white}{[Barak]}
\colorbox{bananamania!11.1815864301!white}{ruled}
\colorbox{applegreen!0.871589869068!white}{out}
\colorbox{bananamania!4.09617424623!white}{an}
\colorbox{bananamania!8.37657592347!white}{early}
\colorbox{bananamania!3.8814177343!white}{resumption}
\colorbox{bananamania!7.93824097303!white}{of}
\colorbox{applegreen!0.97218531833!white}{peace}
\colorbox{bananamania!0.976493408942!white}{talks}
\colorbox{bananamania!23.8144594564!white}{,}
\colorbox{bananamania!13.8010666781!white}{even}
\colorbox{bananamania!6.18777300531!white}{with}
\colorbox{applegreen!4.43457987349!white}{the}
\colorbox{bananamania!8.95518439565!white}{United}
\colorbox{bananamania!3.64823445038!white}{States}
\colorbox{bananamania!1.40967476624!white}{acting}
\colorbox{bananamania!4.82904918825!white}{as}
\colorbox{bananamania!1.06909953263!white}{intermediary}
\colorbox{applegreen!68.340894019!white}{.}
\colorbox{applegreen!52.2952979856!white}{(Eve)}
\colorbox{bananamania!1.83331386184!white}{(Conette)}
\colorbox{bananamania!13.418693561!white}{reports}
\colorbox{bananamania!11.0120549671!white}{from}
\colorbox{bananamania!0.269411330331!white}{Jerusalem}
\colorbox{bananamania!89.3007920613!white}{.}
\colorbox{bananamania!27.705168099!white}{Defending}
\colorbox{bananamania!100.0!white}{what}
| FEMALE
}

2-\quad\texttt{
\colorbox{applegreen!14.2479381516!white}{Once}
\colorbox{applegreen!6.35194055823!white}{again}
\colorbox{applegreen!14.1747241817!white}{there}
\colorbox{applegreen!0.670692755924!white}{'ll}
\colorbox{applegreen!3.936598978!white}{be}
\colorbox{applegreen!0.477965646788!white}{two}
\colorbox{applegreen!8.07998287318!white}{presidential}
\colorbox{applegreen!3.67119431242!white}{candidates}
\colorbox{bananamania!4.91004557174!white}{missing}
\colorbox{applegreen!6.25938197972!white}{from}
\colorbox{bananamania!0.266128794298!white}{the}
\colorbox{applegreen!3.96054159486!white}{debate}
\colorbox{bananamania!74.5464120059!white}{.}
\colorbox{bananamania!2.77480335424!white}{Pat}
\colorbox{bananamania!2.19296628967!white}{Buchanan}
\colorbox{applegreen!7.66431859416!white}{hardly}
\colorbox{applegreen!0.934417577512!white}{registers}
\colorbox{applegreen!3.74228802488!white}{on}
\colorbox{bananamania!2.59173595524!white}{the}
\colorbox{applegreen!17.3486322127!white}{political}
\colorbox{bananamania!6.50234089304!white}{radar}
\colorbox{applegreen!4.98422655006!white}{this}
\colorbox{applegreen!8.61130346424!white}{year}
\colorbox{bananamania!14.8763825777!white}{.}
\colorbox{applegreen!8.66752373289!white}{And}
\colorbox{applegreen!11.2168265054!white}{Ralph}
\colorbox{applegreen!3.25050764054!white}{Nader}
\colorbox{bananamania!3.78571347366!white}{,}
\colorbox{applegreen!4.32621342788!white}{who}
\colorbox{applegreen!2.40986316132!white}{may}
\colorbox{bananamania!7.83345235715!white}{make}
\colorbox{bananamania!5.66448808153!white}{the}
\colorbox{applegreen!1.04475839656!white}{difference}
\colorbox{bananamania!6.20199874749!white}{between}
\colorbox{applegreen!3.36912215445!white}{a}
\colorbox{applegreen!5.34359673028!white}{[Gore]}
\colorbox{bananamania!12.4172336304!white}{or}
\colorbox{applegreen!5.37474250487!white}{[Bush]}
\colorbox{applegreen!12.633053026!white}{win}
\colorbox{bananamania!9.82013000145!white}{in}
\colorbox{applegreen!12.7879953026!white}{several}
\colorbox{bananamania!4.07304152936!white}{places}
\colorbox{bananamania!100.0!white}{.}
\colorbox{bananamania!1.47514436514!white}{(ABC)}
\colorbox{applegreen!23.234355936!white}{('s)}
\colorbox{bananamania!70.6246034822!white}{(Linda)}
\colorbox{applegreen!7.49528031075!white}{(Douglas)}
\colorbox{bananamania!52.2593623509!white}{was}
\colorbox{bananamania!22.3119853706!white}{with}
| MALE
} \\
\bottomrule

    \end{tabular}
    }
    \caption{Addition interpretation examples with Transformer.}
    \label{tab:add-exp-transformer}
\end{table*}

\end{document}